%% file: main.tex
\documentclass[sigconf]{acmart}

\AtBeginDocument{%
  \providecommand\BibTeX{{%
    \normalfont B\kern-0.5em{\scshape i\kern-0.25em b}\kern-0.8em\TeX}}}

\setcopyright{acmcopyright}
\copyrightyear{2018}
\acmYear{2018}
\acmDOI{10.1145/1122445.1122456}

\acmConference[PACT '22]{PACT '22:  International Conference on
Parallel Architectures and Compilation Techniques (PACT)}{October 10--12, 2012}{Chicago,IL}
\acmBooktitle{PACT '22: International Conference on
Parallel Architectures and Compilation Techniques (PACT), October 10--12, 2018, Chicago, IL}
\acmPrice{15.00}
\acmISBN{978-1-4503-XXXX-X/18/06}

\acmDOI{} 
\startPage{1}

\setcopyright{none}

\usepackage{amsmath,amsfonts}
\usepackage{algorithmic}
\usepackage{soul}
\usepackage{graphicx}
\usepackage{dblfloatfix}
\usepackage{textcomp}
\usepackage{xcolor}
\def\BibTeX{{\rm B\kern-.05em{\sc i\kern-.025em b}\kern-.08em
    T\kern-.1667em\lower.7ex\hbox{E}\kern-.125emX}}

\definecolor{arsenic}{rgb}{0.23, 0.27, 0.80}

\newcommand{\name}{\texttt{BenchPress}}

\begin{document}

\title[Short Title]{BenchDirect: A Directed Language Model for Compiler Benchmarks}\thanks{Our source code is publicly available at \url{https://github.com/fivosts/BenchPress}}\thanks{This work was supported by the Engineering and Physical Sciences Research Council (grant EP/L01503X/1), EPSRC Centre for Doctoral Training in Pervasive Parallelism at the University of Edinburgh, School of Informatics. This work was supported by the Royal Academy of Engineering under the Research Fellowship scheme.}


\author{Foivos Tsimpourlas}
\affiliation{
  \institution{Meta AI Research}
}
\affiliation{
  \institution{University of Edinburgh}            
}

\email{F.Tsimpourlas@sms.ed.ac.uk}

\author{Pavlos Petoumenos}
\affiliation{
  \institution{University of Manchester}            
}
\email{pavlos.petoumenos@manchester.ac.uk}

\author{Min Xu}
\affiliation{
  \institution{Meta AI Research}
}
\email{m1n@fb.com}

\author{Chris Cummins}
\affiliation{
  \institution{Meta AI Research}
}
\email{cummins@fb.com}

\author{Kim Hazelwood}
\affiliation{
  \institution{Meta AI Research}
}
\email{kimhazelwood@fb.com}

\author{Ajitha Rajan}
\affiliation{
  \institution{University of Edinburgh}
}
\email{arajan@inf.ed.ac.uk}

\author{Hugh Leather}
\affiliation{
  \institution{Meta AI Research}
}
\email{hleather@fb.com}

\begin{abstract}
\input{00-Abstract}
\end{abstract}

\maketitle

\section{Introduction}
\label{sec:intro}
\input{01-Introduction}

 \section{Motivation}
 \label{sec:motivation}
\input{02-Motivation}

\section{Approach}
\label{sec:approach}
\input{03-Approach}

\section{Experimental Setup}
\label{sec:experiment}
\input{04-Experiment}

 \section{Results and Analysis}
 \label{sec:results}
\input{05-Results}

\section{Conclusion}
\label{sec:conclusion}
\input{08-Conclusion}

 \section{Related Work}
 \label{sec:background}
\input{06-Background}


\clearpage 
\bibliography{References}



\end{document}

%% file: 00-Abstract.tex
The exponential increase of hardware-software complexity has made it impossible for compiler engineers to find the right optimization heuristics manually. Predictive models have been shown to find near optimal heuristics with little human effort but they are limited by a severe lack of diverse benchmarks to train on. Generative AI has been used by researchers to synthesize benchmarks into existing datasets. However, the synthetic programs are short, exceedingly simple and lacking diversity in their features.

We develop \name, the first ML compiler benchmark generator that can be directed within source code feature representations. \name\ synthesizes executable functions by infilling code that conditions on the program's left and right context. \name\ uses active learning to introduce new benchmarks with unseen features into the dataset of Grewe's et al. CPU vs GPU heuristic, improving its acquired performance by 50\%. \name\ targets features that has been impossible for other synthesizers to reach. In 3 feature spaces, we outperform human-written code from \texttt{GitHub}, \texttt{CLgen}, \texttt{CLSmith} and the \texttt{SRCIROR} mutator in targeting the features of Rodinia benchmarks.

\name\ steers generation with beam search over a feature-agnostic language model. We improve this with \texttt{BenchDirect} which utilizes a directed LM that infills programs by jointly observing source code context and the compiler features that are targeted. \texttt{BenchDirect} achieves up to 36\% better accuracy in targeting the features of Rodinia benchmarks, it is $1.8\times$ more likely to give an exact match and it speeds up execution time by up to 72\% compared to \name. Both our models produce code that is difficult to distinguish from human-written code. We conduct a Turing test which shows our models' synthetic benchmarks are labelled as `human-written' as often as human-written code from \texttt{GitHub}.

%% file: 01-Introduction.tex

\begin{figure}
\centering
    \includegraphics[scale=0.25]{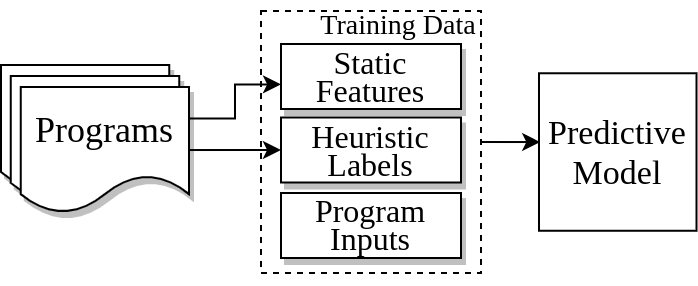}
    \caption{Training pipeline of a predictive model.}
    \label{fig:predictive_model}
\end{figure}

Predictive modeling for compiler optimisation heuristics has been shown to outperform human experts and reduce development time in previous studies~\cite{clgen, Anghabench, tsimpourlas, tsimpourlas_2}. Predictive models learn such heuristics by training on source-level benchmarks or on static code features extracted at the (1) syntax level by traversing their Astract Syntax Tree (AST) or (2) Intermediate Representation (IR) with the help of compiler passes, as shown in Figure~\ref{fig:predictive_model}. However, predictive modeling's effectiveness is restricted by an acute shortage of benchmarks, both in quantity and feature diversity~\cite{8357388, Anghabench, benchpress}, degrading their performance.

There have been some recent generative approaches that leverage the rise of deep learning and language modeling to mitigate this shortage by automatically generating synthetic programs to enhance existing human-written benchmarks~\cite{clgen, clsmith, Anghabench}. While they could provide elegant solutions to improve training data for predictive models, these synthetic benchmarks seem to be short, repetitive with little new features compared to existing benchmarks~\cite{goens}. To generate programs, they either use static programming language specifications with fuzzing or sample programs from learnt distributions, e.g., machine learning algorithms. Their common characteristic is that they generate random benchmarks that are likely to conform to the language's grammar but they are highly unlikely to synthesize benchmarks that are both human-likely and are not already included in existing datasets. What is needed is a systematic method to search for missing programs whose features would be likely to improve the performance of trained downstream tasks. We aim to address this with \name, a targeted benchmark generator, that can generate compiler benchmarks with a desired set of features. In this work, we focus on generating OpenCL benchmarks, as predictive models for heterogeneous systems is a rapidly advancing field and training examples for them are very sparse.

We develop \name~\cite{benchpress_url}, a BERT-based OpenCL benchmark generator~\cite{bert, opencl} that targets and synthesizes benchmarks in desired parts of the feature space.
We use active learning to choose parts of the feature space and beam search to steer \name's generated samples towards the requested features. We train \name\ with OpenCL code samples that we collect by mining BigQuery~\cite{bigquery} and \texttt{GitHub} directly using its API~\cite{github}. We support composite data types and calls to user-defined functions in our dataset and benchmark generation. \name\ is a bidirectional generative model and learns to generate code in any part of a sequence by jointly considering left and right context. We achieve this with a new learnt token, the \texttt{[HOLE]}, which hides a sequence from the input, whose length is unknown to \name\ during training. \name\ learns to fill \texttt{[HOLE]} by iteratively predicting an arbitrary number tokens that are likely to lead to a compiling function. We further develop \texttt{BenchDirect}, an extension of \name\ with a synthesizer conditioned on the features of the complete function. At inference time, this allows us to fill each \texttt{[HOLE]} with code that is more likely to bring us closer to the requested features.

\name\ outperforms \texttt{CLgen} in the task of undirected program generation from a fixed input feed, generating $10\times$ more unique OpenCL kernels that are $7.5\times$ longer on average, with a compilation rate of 86\% compared to \texttt{CLgen}'s 2.33\%. \name\ strongly outperforms benchmark synthesizers \texttt{CLgen}, \texttt{CLSmith}~\cite{csmith, clsmith}, and human written code from \texttt{GitHub} in reaching close to the features of Rodinia benchmarks, developed by compiler experts. The extended synthesizer, by directly filling holes with code that is useful for reaching the targeted features, makes this process 6\% up to 72\% faster, 6\% up to 36\% more accurate and $1.8\times$ more likely to perfectly reach these features. Finally, \name\ uses active learning, specifically query by committee~\cite{qbc}, to search the feature space and find missing features to improve Grewe's et al.~\cite{grewe} CPU vs GPU heuristic. Enhancing the heuristic's dataset with \name's benchmarks improves the heuristic's speedup relative to the optimal static decision by 50\%, increasing it from 4\% to 6\%, when the maximum possible speedup for this task is 12\%.

In this paper, we present the following contributions:

\begin{enumerate}
    \item We are the first to develop a feature-space agnostic, directed code generator towards desired program features.
    \item We develop an automated approach to rank the feature space of downstream tasks with active learning.
    \item We enable bidirectional source code generation by inserting \texttt{[HOLE]} tokens in any part of a sequence.
\end{enumerate}

\subsection{New Contributions}

The contributions of this study, different from our previous work, are summarized as follows:

\begin{enumerate}
    \item We develop \texttt{BenchDirect}, the first bi-directional language model for code infilling that is directed in compiler feature spaces. Compared to \name's language model's random benchmark generation, \texttt{BenchDirect} jointly conditions on code context and target features to generate directly candidates that satisfy them. We conduct an extensive evaluation between \name\ and \texttt{BenchDirect} and we show the latter develops up to 36\% better accuracy in targeting the features of Rodinia benchmarks across 3 feature spaces, while at the same time it requires up to 72\% less time.
    \item We evaluate the human-likeness of \name's,\\ \texttt{BenchDirect}'s, \texttt{CLgen}'s and \texttt{CLSmith}'s benchmarks as a means to measure their quality. We find benchmarks generated by \name\ and \texttt{BenchDirect} to be `human-written' labelled as often as code from \texttt{GitHub} from participants in a Turing test.
\end{enumerate}

%% file: 02-Motivation.tex
\begin{figure}[ht]
\centering
    \includegraphics[scale=0.19]{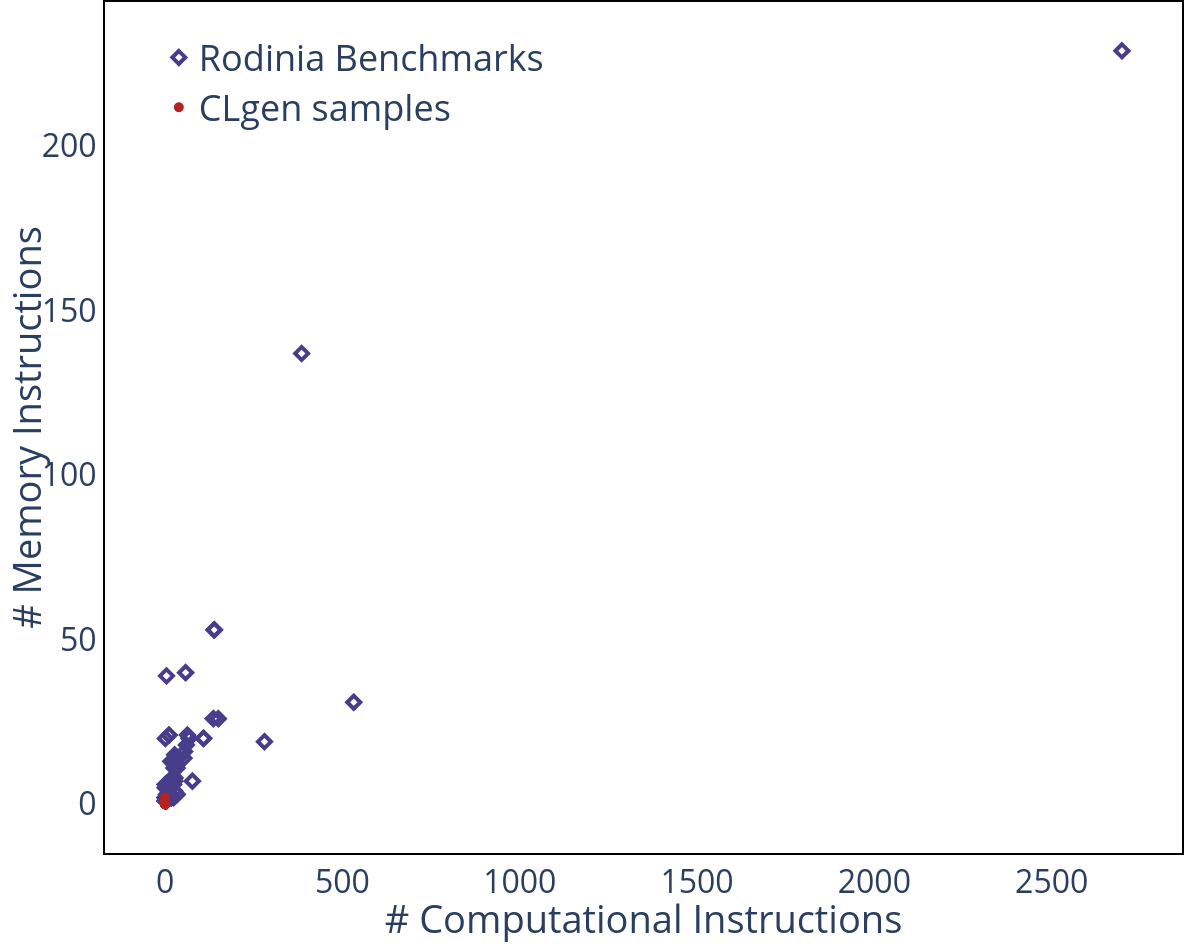}
    \caption{\# Memory operations and \# computational instructions for \textbf{(a)} Rodinia benchmarks in purple diamonds and \textbf{(b)} \texttt{CLgen}'s samples in red dots. Generating samples with missing features is vital for predictive modeling's performance.}
    \label{fig:motivating_example}
\end{figure}

Figure~\ref{fig:motivating_example} shows a two-dimensional slice of the Grewe's et al.~\cite{grewe} feature space: number of computational instructions vs number of memory instructions. Figure~\ref{fig:motivating_example} also shows how the OpenCL benchmarks found in the Rodinia suite map into this plane, represented as purple diamonds. We find much of this two dimensional space is uncovered. 54 of the 58 Rodinia examples cluster in the lower left corner, the rest of the space having only four examples. Any optimization decision for programs in this area of the space would not be accurate due to lack of representative examples. 

\texttt{CLgen} attempted to address this problem by automatically generating more training examples. However, the generated kernels lacked feature diversity and provided even poorer coverage of the feature space. Figure~\ref{fig:motivating_example} represents their position in the 2D space as red dots. Almost all of them are concentrated in a corner covering a small percentage of the feature space. While \texttt{CLgen} can generate hundreds of millions of unique kernels, almost all of them will fail to compile. As the probability of having at least one illegal token in the kernel body increases with the number of tokens, only tiny kernels are valid. In our experiments in Section~\ref{sec:results}, the longest compiling \texttt{CLgen} kernel had 8 lines and 102 tokens. Given the small number of tokens in valid kernels, there is a high degree of repetitiveness in the generated corpus, not only in terms of features but also in terms of structure and functionality. As a result, this approach is not well suited to augmenting the training set with diverse feature benchmarks. There is a compelling need to generate training points for uncovered regions of the feature space and we attempt to address this need with \name. In the following Sections, we discuss our approach and evaluation of \name, comparing it to the existing state-of-the art for feature space coverage.



%% file: 03-Approach.tex
We present \name, a deep learning model for directed compiler benchmark generation. \name\ is the first directed synthesizer for compiling functions with features targeted by a user or a downstream task. \name\ consists of an undirected language model that is trained on source code and a beam search sampler that steers its generation. Given a downstream task, our model uses active learning to search desired features and direct its program generation towards areas of high importance for the task. We further extend \name's underlying language model into a directed synthesizer by encoding compiler features into the model's training process. This enables token generation to attend directly on the targeted features, significantly optimising steerable synthesis. We name this architecture \texttt{BenchDirect}.

\name\ and \texttt{BenchDirect} share a BERT-based language model~\cite{bert}, which we transform into a generative model. There are two key features in our language model that enable directed, bi-directional program generation. First, we develop a new token, namely, the \texttt{[HOLE]}, and we train \name\ to iteratively fill holes of unknown length at any part of an input sequence by conditioning it on the left and right context of the \texttt{[HOLE]}. As an extension to this, \texttt{BenchDirect}'s language model includes a Transformer-based encoder~\cite{attention} that incorporates target compiler features into token classification. This allows tokens to be selected not only with respect to the input's source code context, but also given the compiler features that are targeted. 

\begin{figure}[ht]
\centering
    \includegraphics[scale=0.2]{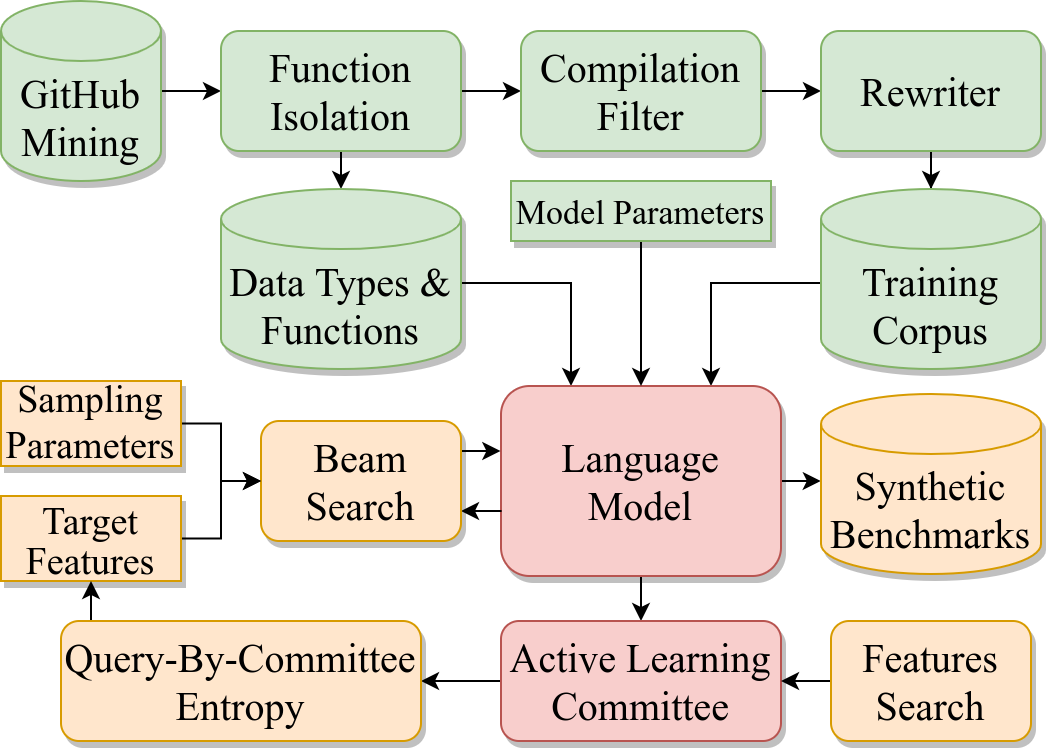}
    \caption{\name's high-level approach.}
    \label{fig:approach}
\end{figure}

Figure \ref{fig:approach} illustrates an overview of our approach. \name\ consists of three main components:

\begin{enumerate}
    \item Learning corpus collection and processing.
    \item Directed source code language modeling.
    \item Feature space search and benchmark generation.
\end{enumerate}

 We discuss each step in the following four subsections. In our last subsection, we discuss \textsc{BenchDirect}'s directed language model, which distinguishes it from our base architecture, \name. Our codebase and experimental data are publicly available~\footnote{https://github.com/fivosts/BenchPress} for researchers to use.

\subsection{Learning Corpus}

Modeling source code accurately requires large amounts of data~\cite{deep_learning} similarly to other deep learning tasks. We develop a tool to collect data from BigQuery's \texttt{GitHub} dataset~\cite{bigquery}. We also use \texttt{GitHub}'s API~\cite{github} and mine directly extra repositories that are not included in BigQuery.

There are a few innovations in how we pre-process the code compared to previous works. First, we inline included header files recursively into source files to resolve type dependencies. Additionally, we automatically extract custom data types (e.g. \texttt{struct}, \texttt{typedef}) and utility functions found in the unprocessed corpus and place them into header files that are accessible throughout \name's pipeline. This way, we resolve most type dependencies by retaining the functionality and semantics of the original, human-written programs. These two steps enable us to increase significantly the amount of compiling kernels we end up with in our training dataset. Second, we isolate kernels into single instances because \name\ is trained on complete functions. From the previous steps, the type dependencies of each kernel are known and we automatically provide them to the compiler, retaining their compilability. Finally, we compile all kernels with Clang and reject those that do not compile.

Next, we re-write identifiers by randomly sampling the alphabet, eliminating spurious naming patterns in the corpus. All kernels are padded to \name's sequence length and kernels that are longer than this are truncated to fit. This helps \name\ train its later indices' positional embeddings more effectively, for which we have less training information compared to earlier indices. 
Finally, we derive a tokenizer by parsing the AST of all source code. We reserve tokens for all OpenCL keywords and all intrinsic OpenCL function name identifiers found in the official OpenCL specifications~\cite{opencl_spec}. We analyze the dataset and tokenize by word the most common function names and custom data type identifiers that we have collected. We encode all literals and infrequently used custom types and functions character by character to avoid exploding the size of the vocabulary. We  define 5 meta tokens: \texttt{[START]}, \texttt{[END]}, \texttt{[PAD]}, \texttt{[HOLE]}, \texttt{[ENDHOLE]}. The derived tokenizer holds in total 2,201 unique tokens.

\subsection{Language Modeling}
\label{subsec:language_modeling}

\name\ is based on BERT~\cite{bert}, a Transformer-based model originally designed for natural language modeling. BERT is trained to predict words that have been randomly hidden by \texttt{[MASK]} tokens. This way BERT learns fitting words with respect to their position in a sequence and also the left and right context, i.e., the text sequence before and after the masked token to be predicted. This type of training helps BERT learn what words mean within a given context, improving downstream tasks that rely on that knowledge. 

While this is a useful property, it is not enough to turn BERT into a generative model. We also want to be able to extend a kernel by inserting an arbitrary number of tokens in arbitrary positions. We could iteratively add a \texttt{[MASK]} token to get one extra token at a time, until we have a full statement. This would be limiting. Each time the new token would be selected based on its probability of completing forming a plausible kernel. Every intermediate kernel in the iterative process would have to be plausible or almost plausible, which is not a general way for augmenting kernels.

Clusters of \texttt{[MASK]} tokens could allow us to insert multiple tokens in each iteration. This is still unsatisfactory. The number of \texttt{[MASK]} tokens in the cluster biases the kind of code that will be generated: if we ask such a generator to produce five tokens, it will give us a five token statement that could be expected to close this gap, not a five token sequence that could be the start of a much longer statement. We could place the left and right context to the edges of a sequence and fill intermediate positions with [MASK] tokens. \name\ could predict a vocabulary or a stop token for a [MASK], allowing for arbitrary sequences. We test this configuration and sample a trained model with a fixed input feed. \name\ is unable to learn the [MASK]s' left and right context conditionally, when many [MASK]s are in a sequence, which leads to zero samples to compile or even resemble reasonable code.

What we do instead is to extend BERT's functionality with a new pair of learnt tokens, the \texttt{[HOLE]} and the \texttt{[ENDHOLE]}. \texttt{[HOLE]} follows the same logic with \texttt{[MASK]}, however the number of tokens that have been hidden behind it is unknown to the model during training. The model only learns to predict the first token of an arbitrarily long missing sequence. At inference-time, we iteratively predict the first token of the remaining sequence and re-insert it just before the \texttt{[HOLE]}. This way \name\ learns to generate arbitrarily large code sequences within any part of a sequence.


\begin{figure}[ht]
\centering
    \includegraphics[scale=0.2]{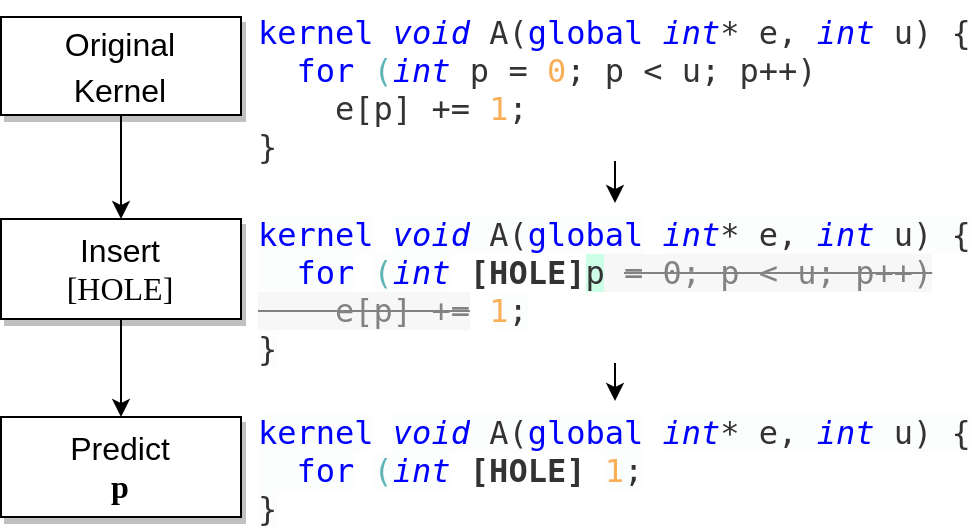}
    \caption{When a \texttt{[HOLE]} is inserted to a kernel at a random index, it hides a random number of tokens, unknown to \name. On this example, \name\ learns to predict the first hidden token, \textbf{\texttt{p}}.}
    \label{fig:insert_hole}
\end{figure}

Figure \ref{fig:insert_hole} shows how a \texttt{[HOLE]} is inserted into a function to create a datapoint. A random starting index and a random length are selected. The choice of index and length are only restricted by a potential overlap of the prospective hidden sequence with any of the other meta token or the maximum hole length that is defined as a training parameter for the architecture as a percentage of each function's length. When the specifications of a hole have been settled, the hidden sequence is discarded. Only the first token of it is kept as the target prediction for that hole. A hole can also represent an empty sequence, i.e. hiding 0 tokens. In this case, the target prediction during training is \texttt{[ENDHOLE]}. The training instances are randomly generated on demand, the entire space of possible instances is too large to be pre-generated. 
In this paper, we only insert 1 hole per training instance for \name\ to learn. Multiple holes could be used during training, but this is not needed during \name's current benchmark generation task.

\subsection{Benchmark Generation}

\begin{figure}[ht]
\centering
    \vspace{-5pt}
    \includegraphics[scale=0.2]{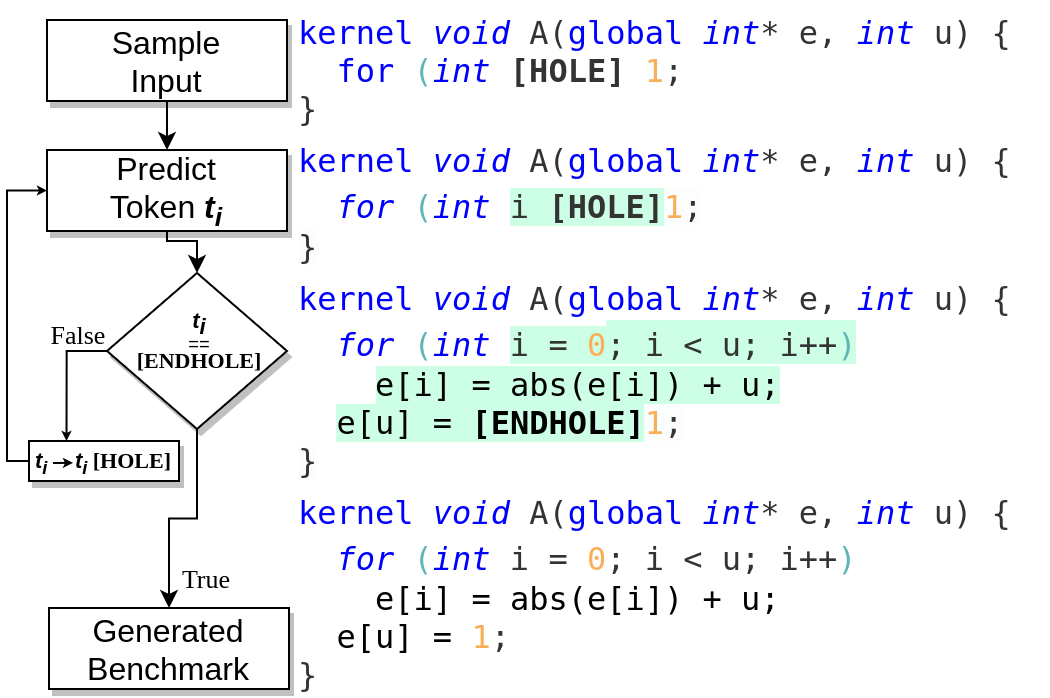}
    \caption{During sampling, \name\ receives an input and predicts iteratively the fitting tokens. \name\ predicts \texttt{[ENDHOLE]} to indicate a \texttt{[HOLE]} is complete.}
    \label{fig:fill_hole}
\end{figure}

\name's synthesizer operates as a generative model with the help of \texttt{[HOLE]} / \texttt{[ENDHOLE]} tokens. It receives an input with 1 or more \texttt{[HOLE]} tokens and returns a completed benchmark. For each \texttt{[HOLE]}, \name\ predicts one token that fits in the sequence at the \texttt{[HOLE]}'s index, with respect to its left and right context. If the predicted token is not \texttt{[ENDHOLE]}, it moves the \texttt{[HOLE]} and all subsequent tokens one position to the right and inserts the predicted token to the initial target index. This intermediate kernel is iteratively provided as an input for the next token prediction and the process is repeated until \name\ predicts \texttt{[ENDHOLE]}. This marks a \texttt{[HOLE]} is complete and the final sample is returned, as shown in Figure \ref{fig:fill_hole}.

On its own, this process only augments kernels given their existing left and right context. In that sense, \name's language model is undirected with respect to the features that are targeted. We make \name\ the first synthesizer to target desired parts of a feature space with beam search sampling. We generate a set of kernels from an empty input, we select the ones closer to the target features and we insert holes to generate new edited kernels iteratively.

Given a target feature vector, \name\ samples a starting, fixed input feed \texttt{`kernel void [HOLE]'} and yields a collection of starting benchmarks. We reject benchmarks that do not compile and for the remaining we measure the Euclidean distance between their feature vectors and the target features. We select the \textit{top-K} candidates that have the shortest distance from the target and we use them as inputs for the next generation. To improve diversity among promoted benchmarks we introduce randomness in the selection of \textit{top-K} candidates: Each \textit{top-K} sample, has a fixed probability $p=0.15$ to be replaced by another random candidate of its generation. \name\ lazily creates multiple different input instances for each selected candidate by placing a random \texttt{[HOLE]} of random length in order to synthesize a new sample. \name\ generates a successive collection of benchmarks, of which \textit{K} compiling ones with the shortest distance from the target again are selected with \textbf{\textit{p}}-randomness and used as inputs. This search continues until a sample achieves a distance of $0$ from the target, or until a threshold of generations (i.e. beam search depth) is exhausted. \name\ returns the closest benchmark to the target's features along with all beam search's intermediate benchmarks that cover the model's traversal of the feature space starting from the origin and ending near the target features. For the benchmark synthesis process, we use categorical sampling with temperature to sample \name's probabilities. The sampling temperature, beam search's width \textit{K} and depth are defined as sampling parameters.

In the worst case, \name's directed program generation is slow, ranging from a few seconds to one hour, as it typically requires thousands of random language model inferences. However, \name\ is the first program synthesizer that can target a set of desired program features. \texttt{BenchDirect} speeds up targeting features significantly as its directed language model requires far less samples per beam search iteration to produce samples close to the target features. Often, \texttt{BenchDirect} can target the feature space within one single inference step from an empty input.

\subsection{Feature Space Search}

A steerable synthesizer allows the generation of benchmarks with desired features. However, the automatic selection of those parts of the feature space that are worth targeting is challenging and depends on the downstream task.

\name\ attempts to solve this by searching the feature space with query by committee~\cite{qbc}, a well-known active learning technique. We implement a committee of (a) 7 NN, (b) 7 k-NN and (c) 7 K-means models. We set their initial state by passively training on a small portion of the downstream task's data. We sample the committee with thousands of random points in the space, we collect the predicted labels and measure the entropy for each sample. The entropy shows the level of uncertainty among the committee about the predicted label of a given point and is defined as:

\begin{equation}
    H = -\sum^{l \epsilon L}(p(l) * \log(p(l)))
\end{equation}

where $L$ is the set of all predicted labels and $p(l)$ the probability of label $l$ in the committee's prediction set for a given input. The highest entropy point is an important feature vector to target and \name\ steers benchmark generation towards it with the approach explained in 3.3. We collect the labels of generated benchmarks and we train incrementally the committee with them. Then, we sample it to find the next highest entropy point. We continue this process until we saturate the feature space. \name's committee is agnostic to the downstream task or the feature space and its I/O dimensions are hyper-parameters selected with respect to the task's feature and prediction dimensions.

\subsection{Directed Language Modeling}

\name's synthesizer presented thus far is feature agnostic. This language model infills source code given the input context left and right of the \texttt{[HOLE]}. \name\ is only able to steer program generation through a costly beam search on the model's output: we generate a large number of random code candidates and we feed those that are closer to the target features back into the model's input with new holes for further edits. Given \name's language model is undirected, it often needs hundreds of thousands of code candidates to increase the chance of finding a few with the right features. This is inefficient and unsustainable on complex compiler tasks.

Instead of randomly trying to fill the space with new benchmarks to get closer to the target features, an approach to target them directly during synthesis is needed. At the best case, this would help collect a benchmark with the right features in a single inference. To this end, we develop \texttt{BenchDirect}, a steerable program generator that extends \name's undirected language model into a directed one. Along with the masked source code input, \texttt{BenchDirect} also encodes its compiler features before masking. Its classification head selects tokens to fill a \texttt{[HOLE]} by jointly observing the code context and the encoded features. This leads to selecting tokens that are likely to generate a kernel that is (a) compiling, similarly to \name, but also (b) matching the target features provided in the input. Even if \texttt{BenchDirect} cannot target a set of features within a few attempts, combined with our beam search sampling, further edits can be made efficiently until it does.

\begin{figure}
\centering
    \includegraphics[scale=0.65]{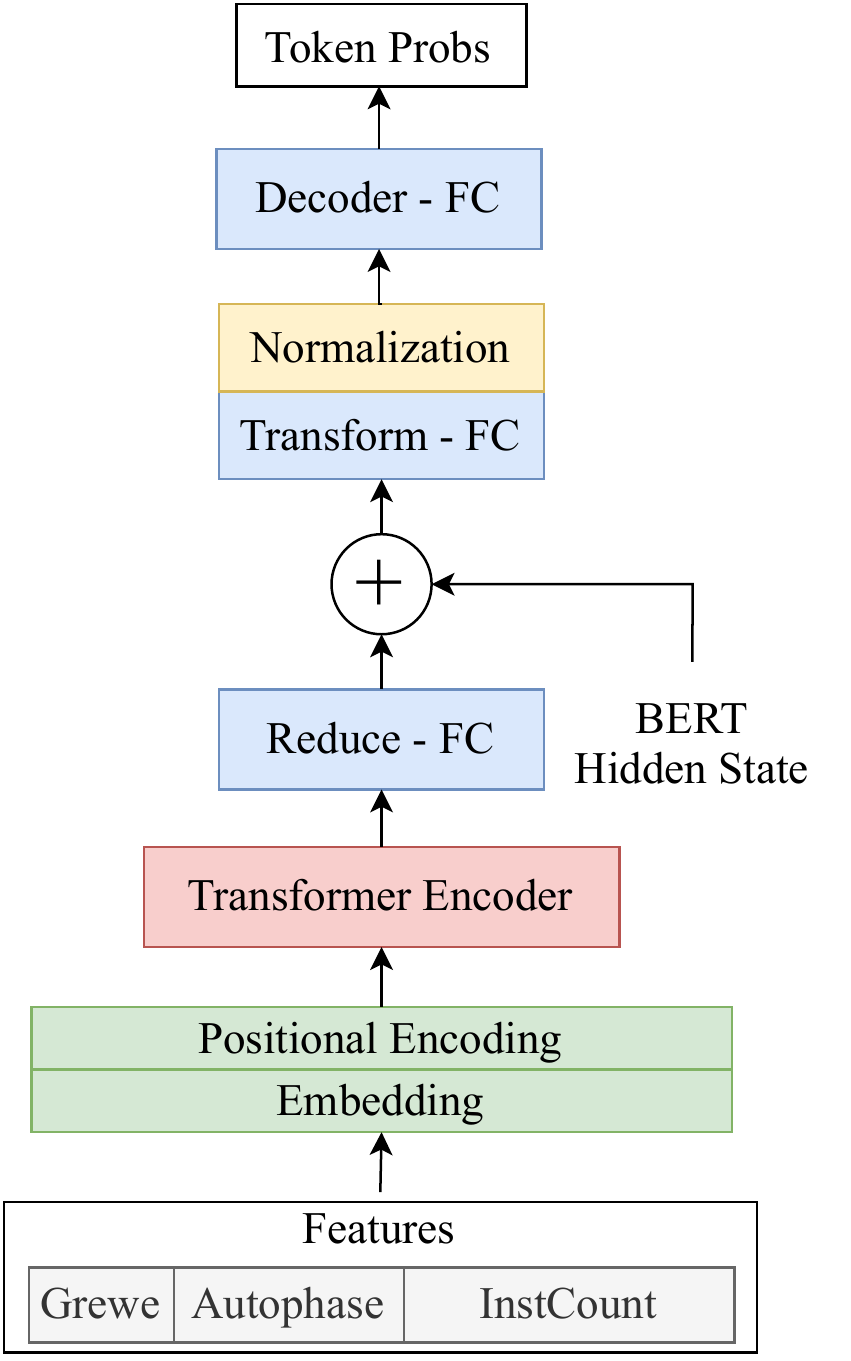}
    \caption{\textsc{BenchDirect}'s directed language model design.}
    \label{fig:feat_transformer}
\end{figure}

\texttt{BenchDirect}'s extended feature encoder is based on Transformer~\cite{attention} and is shown in Figure~\ref{fig:feat_transformer}. We encode a vector of numerical compiler features using an embedded layer with positional encoding followed by a Transformer-Encoder. We reduce the dimensions of the Transformer's output using a Fully Connected layer to match BERT language model's hidden state representation of its input source code. Both hidden states are concatenated and fed to a Fully Connected layer with GELU~\cite{gelu} activation to extract correlated features. Finally, a Decoding Fully Connected layer projects the joint hidden state into the vocabulary space. The feature encoder's input consists of 134 positions divided into three fixed segments. Each represents one feature space used in our evaluation: (a) 8 positions for Grewe's et al. features, (b) 56 for Autophase and (c) 70 for InstCount features. \texttt{BenchDirect} can support multiple spaces and it only needs to be trained once to direct benchmark synthesis on any of them. To steer generation in a new feature space, we simply need to extend a new segment in the Transformer-Encoder's input and apply fine-tuning using the new space's feature extractor to collect data from our training corpus.

\texttt{BenchDirect} is trained with the same approach described in Subsection~\ref{subsec:language_modeling}. We sample randomly one OpenCL kernel and introduce a \texttt{[HOLE]} to provide it to the language model's input. The model learns to predict the first token of the hidden sequence using cross categorical entropy loss function. Introducing compiler features in training is the distinction to this process. When one OpenCL kernel is sampled, its compiler features are also collected. The model receives a pair of inputs, $(src_i, fv)$ and one output $token_i$, where $i$ is the index at which the \texttt{[HOLE]} is located.

It is important to note that we do not feed the feature vectors of all three feature spaces to the encoder at the same time. Instead, we uniformly select one, we set its values to the respective segment of the encoder's input and we \texttt{[PAD]} all other positions such that gradients are not applied. Over training time, the model observes datapoints from all feature spaces for every kernel. Padding all feature spaces but one allows the trained model to learn how to direct synthesis to each one of them independently. Providing vectors from all spaces as one datapoint would possibly allow the model to learn correlations between them but this is not useful to us. What is more, directed synthesis on one of the feature spaces would be impossible. The model would have been trained to observe all three feature vectors for one given source code input. This means we would have to know the mapping function among all feature spaces to translate a target feature vector to all supported ones for the encoder's input. Instead, keeping one feature space per datapoint leads to the encoder's weights to be tuned accordingly to perform accurately on all spaces separately. Parts of the network (e.g. the FC layers) are jointly trained to optimise all feature spaces encoding. Other parts, such as the $(Q, K, V)$ matrices are grouped in vectors, one for each index separately, and are only trained when their respective positions are not padded. An alternative solution would be to use many Transformer-Encoders, one per feature space, and train each separately. During generation, the appropriate Transformer would be manually selected given the desired feature space. Although this is a valid approach, there is no evidence to suggest it would perform better than one Transformer model large enough to learn all segments separately.

During sampling, \texttt{BenchDirect} receives a source code input and the target features as an input. Given the code context and the \texttt{[HOLE]} position, the model will attempt to select those tokens that will produce a compiling kernel with features as close as possible to the target in that respective feature space. At its best, we hope \texttt{BenchDirect} can receive an empty code input and provide the target benchmark at a single inference step. At the very least, the beam search sampler will go through fewer iterations and fewer inferences per generation compared to \name.

%% file: 04-Experiment.tex
We describe the configurations used in training \name, and the parameters used in evaluation, namely (1) Feature Spaces - we use three different representations of program features, (2) Target Benchmarks - We use Rodinia benchmarks~\cite{rodinia} and their features as the target for synthesis by \name, (3) Comparison to SOTA - we compare \name\ with code synthesizers and human written code in improving Grewe's et al. heuristic model.

\subsection{Platforms}
\label{subsec:platforms}

We train \name\ and conduct all our experiments on two 64-bit systems each having one Intel Xeon E5-2620 16-core CPU, 2x Nvidia GeForce GTX 1080 GPU and 32 Gigabytes of RAM. We use Ubuntu 18.04, PyTorch 1.9.1~\cite{pytorch}, CUDA version 11.4 and Nvidia driver version 510.47.03. We use Clang-10 as \name's compiler and LLVM-10 to compile and execute InstCount and Autophase~\cite{autophase} extracting tools. For compatibility reasons, we are required to use Clang LibTooling from LLVM-6 to execute Grewe's et al.~\cite{grewe} feature extractor.

\subsection{Language Modeling for source code}
\label{subsec:config}

We collect OpenCL code from \texttt{GitHub} and split it into single function instances. We ensure no kernels that come from benchmarks suites used in the evaluation are included in our corpus. We pre-process text, re-write variables and reject OpenCL kernels that do not compile. In total we mine 63,918 OpenCL kernels across 12,860 \texttt{GitHub} repositories and we successfully compile 19,637 of them (31\% compilation rate).


We train \name\ on our OpenCL Corpus for 10M steps with a batch size of 32. For \name's BERT model parameters, we select 2 hidden layers, 12 attention heads. We set intermediate size, hidden size and max position embeddings to 768. We set the maximum length of holes to be 90\% of a kernel's token length, i.e. a hole can hide almost all tokens of a training instance. We optimize the model using Adam optimizer with a learning rate that reaches a maximum of $45x10^{-6}$ after 20,000 warmup steps and decays linearly over the remaining training steps. We train \name's language model to a final loss value of $0.28$.

\subsection{Feature Spaces}

Compiler predictive models use static code features to represent programs and learn optimisation heuristics. A vector of independent characteristics represent a single program. Each of them are typically an integer or float value. Features are extracted at the Syntax level by traversing the AST or at the IR level using the compiler's middle end (e.g. LLVM-IR). A feature space is the collection of all possible program feature vectors.

\name\ is a generative model that can be steered to generate samples for a desired part of the feature space. We evaluate \name\ on three source feature representations we find across the literature, (a) Syntax-level Grewe's et al. features~\cite{grewe}, (b) IR-level LLVM-InstCount~\cite{LLVM} and (c) IR-level Autophase~\cite{autophase}.

Grewe's et al. features are extracted with Clang's LibTooling and used to train their predictive model on the CPU vs GPU task for OpenCL kernels. This feature space holds 8 dimensions. 4 dimensions describe the number of 1) computational, 2) relational, 3) atomic and 4) memory access instructions. The feature space also counts the different type of memory instructions, local memory or coalesced. Finally, the computational to memory and coalesced to memory ratios are defined.

InstCount is a standard pass provided by LLVM-IR framework and used in Compiler Gym by Cummins et al.~\cite{compiler_gym}. InstCount holds 70 dimensions: 67 dimensions each counting all 67 LLVM-IR instruction types and total number of 1) instructions, 2) basic blocks and 3) functions. Autophase by Huang et al.~\cite{autophase} holds 56 dimensions. While many of the features used in Autophase are shared with InstCount, they introduce new ones such as number of input arguments to PHI Nodes or total number of memory instructions. On the other hand, they do not include the count of some LLVM instructions that are not considered to contribute to a program's representation, e.g. \texttt{CatchPad} instruction.

\subsection{Analysis of \name\ and CLgen language models}
\texttt{CLgen}~\cite{clgen} is the current state of the art in OpenCL benchmark generation. Its synthetic benchmarks improve the accuracy of Grewe's et al. predictive model~\cite{grewe} by $1.27\times$. However, Goens et al.~\cite{goens} perform a case study and show evidence that \texttt{CLgen}'s synthetic benchmarks do not improve the quality of training data and, consequently, performance of predictive models. They show that a predictive model in fact performs worse with synthetic benchmarks as opposed to human written benchmarks or code from \texttt{GitHub}.

This study motivates us to perform an analysis of \name's language model, BERT, with \texttt{CLgen} in the task of undirected program generation. In this first experiment, we reproduce \texttt{CLgen} using the authors' artifacts and we sample it with a fixed input \texttt{`kernel void'} to collect a dataset of unique OpenCL kernels. We use \name\ on the same generative task and sample the model with the same fixed input \texttt{`kernel void [HOLE]'} to obtain another dataset of unique benchmarks. In this experiment we focus on the language model's inference performance. We compare both generative models on their throughput, their ability to create compiling code, feature distribution and code size. In this experiment, we do not direct program generation. \name\ generates compiling kernels in a single inference step.

\subsection{Targeted Benchmark Generation}
\label{subsec:target_bench}

Next, we evaluate \name's ability to steer towards desired program features. We use well-established compiler benchmarks as our reference and target their features within this space. These benchmarks usually perform intensive operations, such as matrix multiplications or FFT analysis, they contain hundreds of computational and memory instructions and are specifically fine-tuned by experts to exercise compilers from different angles. As a result, we believe features in these benchmarks provide a good target to assess performance of \name's ability to target complex features.

We choose target benchmarks within the Rodinia suite~\cite{rodinia, rodinia_download} as it is widely used in the literature~\cite{clgen, Anghabench}. Similar to the training corpus, we collect the suite's source files, we inline header files and dependent OpenCL libraries into them, we split kernels into single source files and reject those that do not compile. In total, we collect 61 target Rodinia benchmarks out of which 58 compile. For the remaining benchmarks, we collect their features using the feature extractors for Grewe's et al., InstCount and Autophase feature spaces~\cite{grewe, LLVM, autophase}. We target the feature vectors of these benchmarks and request \name\ to generate at least one matching benchmark for each. We end up with three collective synthetic benchmark datasets, one for each feature space, that contain code with features matching Rodinia benchmarks. For each Rodinia benchmark's target feature vector, we measure the minimum Euclidean distance to it achieved between \name, code from GitHub, \texttt{CLgen} and \texttt{CLSmith}~\cite{csmith, clsmith}. For \texttt{GitHub}'s and \texttt{CLSmith}'s kernels, we use \texttt{SRCIROR}~\cite{srciror} to apply code mutations exhaustively with beam search.

 To make our experiment more intuitive we use two datasets for \texttt{GitHub}: a) \texttt{GitHub} consisting of all OpenCL kernels we collected and b) \texttt{GitHub-768}, a proper subset of \texttt{GitHub} which contains only the kernels that do not exceed \name's sequence length of 768 tokens. Since \name\ benchmarks' size are restricted to the architecture's sequence length, we feel it is important to make this distinction in order to present a view of \name's actual performance on features that may be unreachable within the current sequence length. For example, it may be impossible to generate 2,000 computational instructions within 768 tokens. For such cases, we believe \texttt{GitHub-768} with its equally restricted sequence length would allow for a fairer comparison. 

For all three feature spaces, we weed out the Rodinia benchmarks that have an exact matching sample (i.e. a Euclidean distance of 0) in \texttt{GitHub-768}. Since we already have matching samples for them, we do not need to target them with \name\ or any other generative model. However, we do not skip benchmarks \sloppy{whose} features exist only in \texttt{GitHub}'s full dataset as we wanted to explore the feasibility of using \name\ to generate a sample with the same features but smaller sequence length. Applying this restriction we end up with 22 Rodinia benchmarks for Grewe's et al., 52 for InstCount and 36 for Autophase feature spaces.

We sample \name\ for a maximum of 50 beam search iterations unless a benchmark matching the target features is produced. We set a workload size of 2048 samples per iteration. Among those of them that compile, our beam search sampler propagates to the next generation the closest 32 candidates, placing new holes into them.




\subsection{Active Learning for Feature Selection}

\name's steerable generation is vital for searching the feature space while also finding useful features to target with active learning. In this experiment, we evaluate \name\ in the downstream task of training the predictive model proposed by Grewe et al.~\cite{grewe}, a well-tested problem used by many baseline models.

Grewe et al. train a decision tree model to predict the optimal device to execute a benchmark, choosing between a CPU and a GPU. They measure their model's performance as speedup achieved with using the predicted device for execution versus statically executing all benchmarks on the GPU. To train the predictive model, they use OpenCL benchmarks from 7 well-known benchmarks suites~\cite{clgen, grewe}. In this experiment, we reproduce Grewe's et al. heuristic using their artifact and we also retrain it with datasets enriched with executable benchmarks from \name\ using active learning and passive learning (i.e. targeting random parts of the feature space instead of searching it), \texttt{CLgen} and \texttt{GitHub}. We measure the speedup over static mapping for each of them.

To collect our evaluated datasets, we execute OpenCL benchmarks with \texttt{CLDrive}~\cite{clgen} by Cummins et al. \texttt{CLDrive} automatically generates inputs and drives kernels to the hardware. It measures the execution time per device across thousands of runs and it rejects kernels that produce runtime errors, do not modify any of the inputs (no output) or modify them differently for each run (not deterministic). For (a) the 7 human-written benchmarks suites, (b) \name, (c) \texttt{CLgen} and (d) \texttt{GitHub}, we execute their kernel on \texttt{CLDrive} using a range of different \textit{local} and \textit{global size} configurations. We label each instance with the fastest measured device (the CPU or the GPU), in the same way Cummins et al.~\cite{clgen} and Grewe et al.~\cite{grewe} performed their evaluation.

\subsection{Directed Language Modeling}

\name\ develops strong performance compared to state of the art program synthesizers and its benchmarks outperform even human-written benchmarks from \texttt{GitHub} in two tasks, (a) targeting the features of Rodinia benchmarks and (b) improving the accuracy of a compiler heuristic model. However, its undirected language model requires up to hundreds of thousands of inferences for its beam search sampler to minimize its samples' distance from the target features. This process can be inefficient, which we strive to address with a directed language model, namely \texttt{BenchDirect}.

We repeat the experiment of Section~\ref{subsec:target_bench} to evaluate \texttt{BenchDirect}'s accuracy and execution time in targeting the features of Rodinia benchmarks compared to \name. We target the features of Rodinia benchmarks in all three feature spaces for a range of different workload sizes: 32, 64, 128, 256, 512, 1024 and 2048. A large workload size leads to a significant time overhead but is required to ensure high accuracy for \name's undirected language model. This may not be the case for \texttt{BenchDirect}'s directed synthesizer, speeding up directed generation without compensating on its accuracy. In this experiment, we explore how this parameter affects accuracy and total execution time for both models.

We re-train \name\ and \texttt{BenchDirect} for 8M steps to a final loss of 0.14 using the same BERT hyper-parameters described in Section~\ref{subsec:config}, except for their max position embeddings which we set to 512 instead of 768 to reduce training time. For \texttt{BenchDirect}'s Transformer-Encoder, we set an embedding size of 512, 4 attention heads, 2 hidden layers and we set its Fully Connected layers to 1024 features. During sampling, we set the threshold of maximum beam search iterations to 5. Reducing the models' sequence length to 512 and the sampler's iteration threshold to 5 leads to a performance reduction compared to \name's accuracy in Section~\ref{subsec:target_bench}. However, it saves valuable compute time. Both \name\ and \texttt{BenchDirect} are restricted by this reduction, therefore the validity of this comparative study's results is not hurt.

\subsection{Human Likeness of Generated Code}

A great challenge for neural synthesizers is to produce programs that are human likely, that is following basic structural and syntactical form that makes them easy for humans to read and understand. The human likeness of a synthetic program reflects its quality and efficiency in the functionality it serves. To this end, we conduct a case study to measure the likeness of \name's generated benchmarks to human-written code. We devise a double blind Turing test in which we show to human participants random samples from \name, \texttt{BenchDirect}, \texttt{CLgen}, \texttt{CLSmith} and also human-written code from \texttt{GitHub}. They are shown randomly selected benchmarks from the stored datasets and are asked to label them as human or AI-written. We release our Turing test publicly available in the form of a web application\footnote{https://humanorai.co.uk}.

%% file: 05-Results.tex
\begin{table*}[]
\begin{tabular}{lllllll}
           & \begin{tabular}[c]{@{}l@{}}\# unique\\ benchmarks\end{tabular} & \begin{tabular}[c]{@{}l@{}}\# compiling\\ benchmarks\end{tabular} & \begin{tabular}[c]{@{}l@{}}compilation\\ rate\end{tabular} & \begin{tabular}[c]{@{}l@{}}max\\ tokens\end{tabular} & \begin{tabular}[c]{@{}l@{}}max inst\\ (LLVM-IR)\end{tabular} & \begin{tabular}[c]{@{}l@{}}time per\\ sample (ms)\end{tabular} \\\hline
BenchPress & 190,460 & 142,607 & 86\% & 750 & 161 & 162 \\
\texttt{CLgen}      & 1,564,011 & 13,035 & 2.33\% & 102 & 32 & 103
\end{tabular}
\caption{Throughput comparison between \name\ and \texttt{CLgen} on generated OpenCL benchmarks when \name\ does not use feature-directed program generation.}
\label{tab:clgen_vs_bp_overview}
\end{table*}

In this section, we show our experiments' results and compare \name\ with state of the art techniques in OpenCL benchmark synthesis. We present case studies of (a) \name's throughput as a generative model compared to \texttt{CLgen}, (b) its ability to steer benchmark generation towards desired features and (c) its performance in searching the feature space to enhance a downstream task's performance.

\subsection{Analysis of \name\ and \texttt{CLgen} language models}
We perform an analysis of \name\ and \texttt{CLgen} as language models and compare them in generating a collection of benchmarks from a fixed input feed, \texttt{`kernel void [HOLE]'} and \texttt{`kernel void'} respectively. We compare the two approaches measuring (a) the generative models' throughput and (b) the quality of their generated benchmarks in terms of code size and features. In this experiment, we do not use any directed search or iterative approach for \name's generation. We perform this evaluation to measure how BERT, \name's underlying language model, compares with \texttt{CLgen} as a generative model. Table~\ref{tab:clgen_vs_bp_overview} presents the aggregate measurements for the generated benchmarks using both approaches.

\begin{figure}[ht]
	\centering
	\setlength{\tabcolsep}{0.01em}
	\begin{tabular}{cc}
        \includegraphics[scale=0.2]{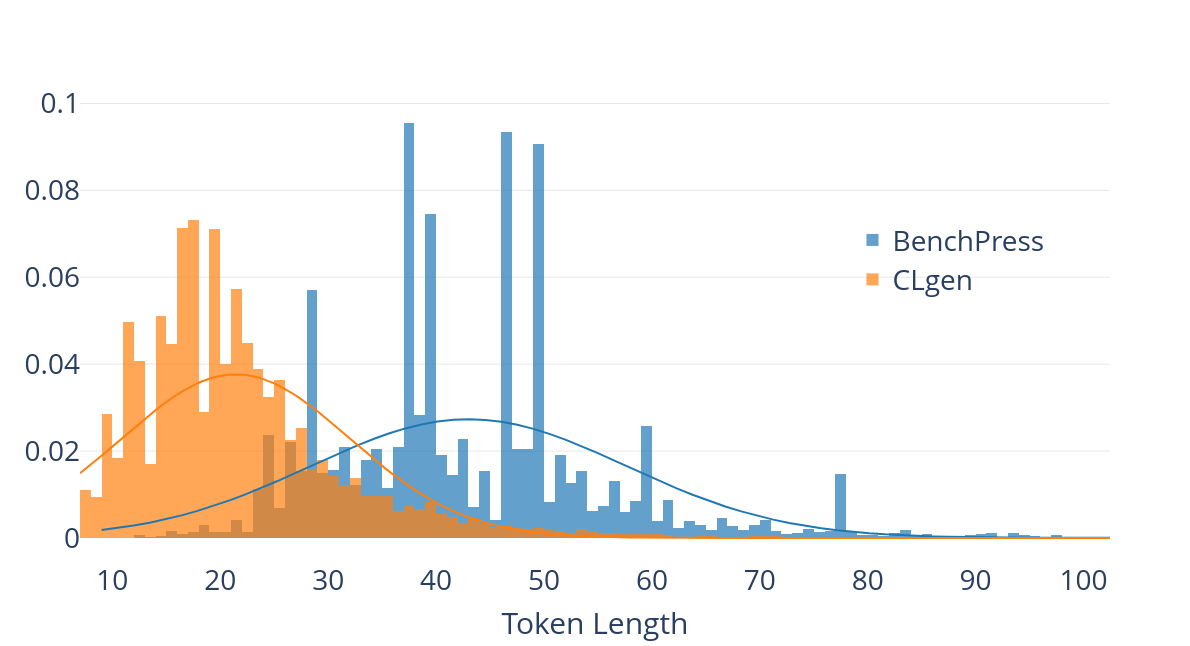}
        \\
        \includegraphics[scale=0.2]{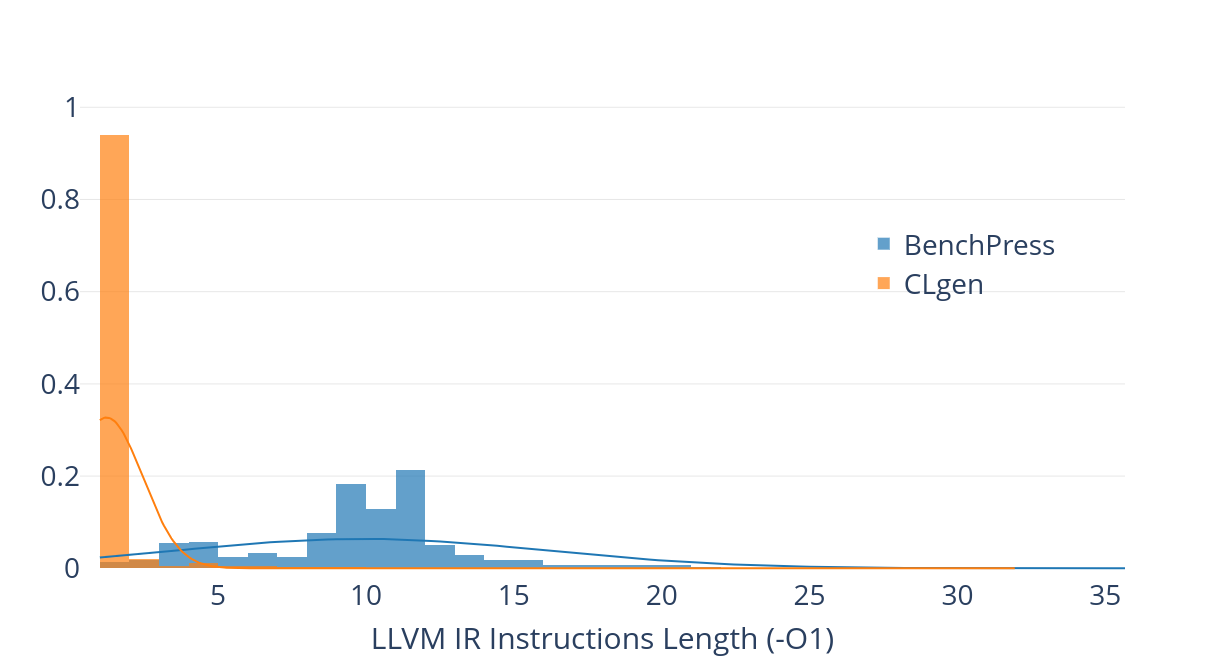}
	\end{tabular}
	\caption{Probability distribution of (a) token length and (b) LLVM-IR Instruction count among \name's and \texttt{CLgen}'s generated benchmarks. \name's benchmarks presented here are generated at a single inference step without iteratively directing program synthesis.}
	\label{fig:size_distribution}
\end{figure}

\begin{figure*}[]
	\centering
	\setlength{\tabcolsep}{0.14em}
	\begin{tabular}{ccc}
        \includegraphics[scale=0.139]{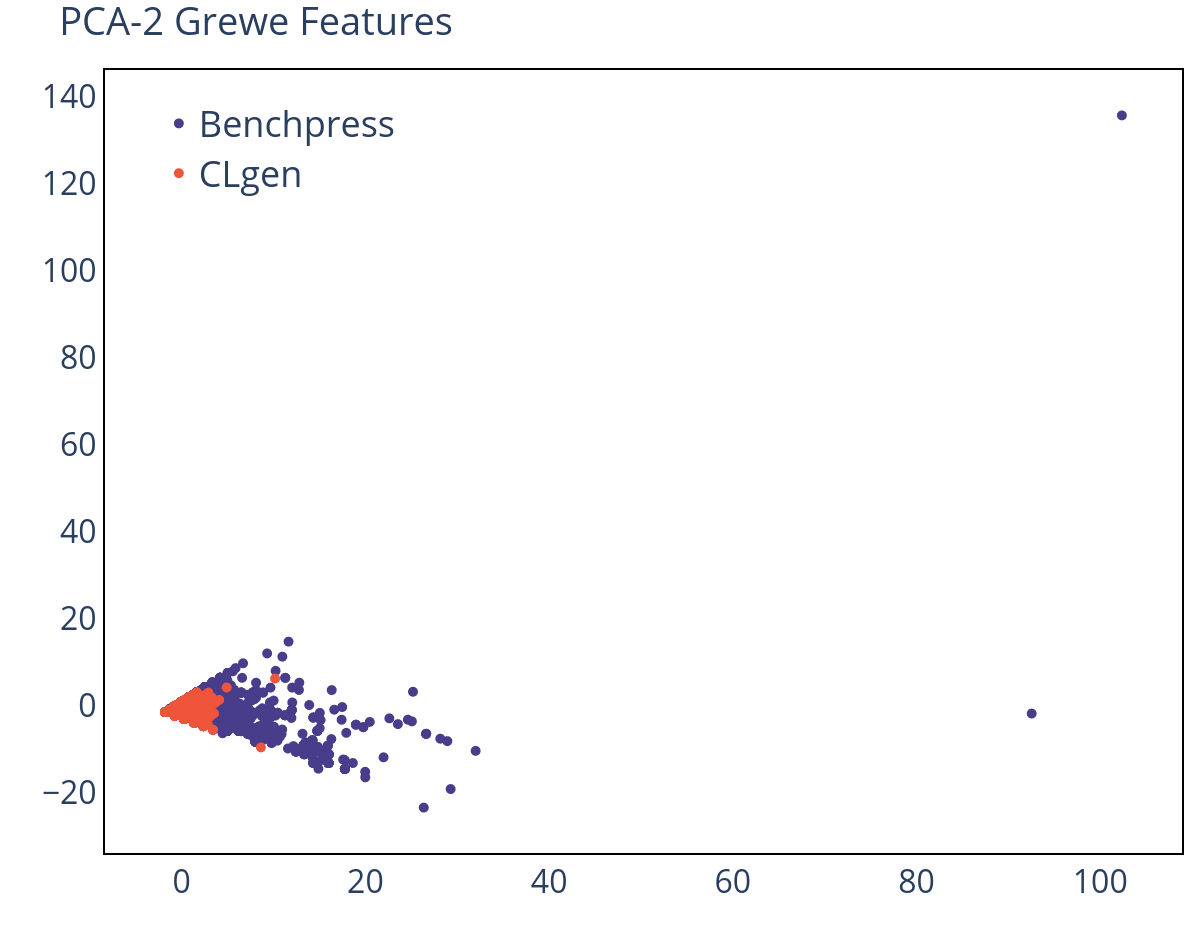}
        &
        \includegraphics[scale=0.139]{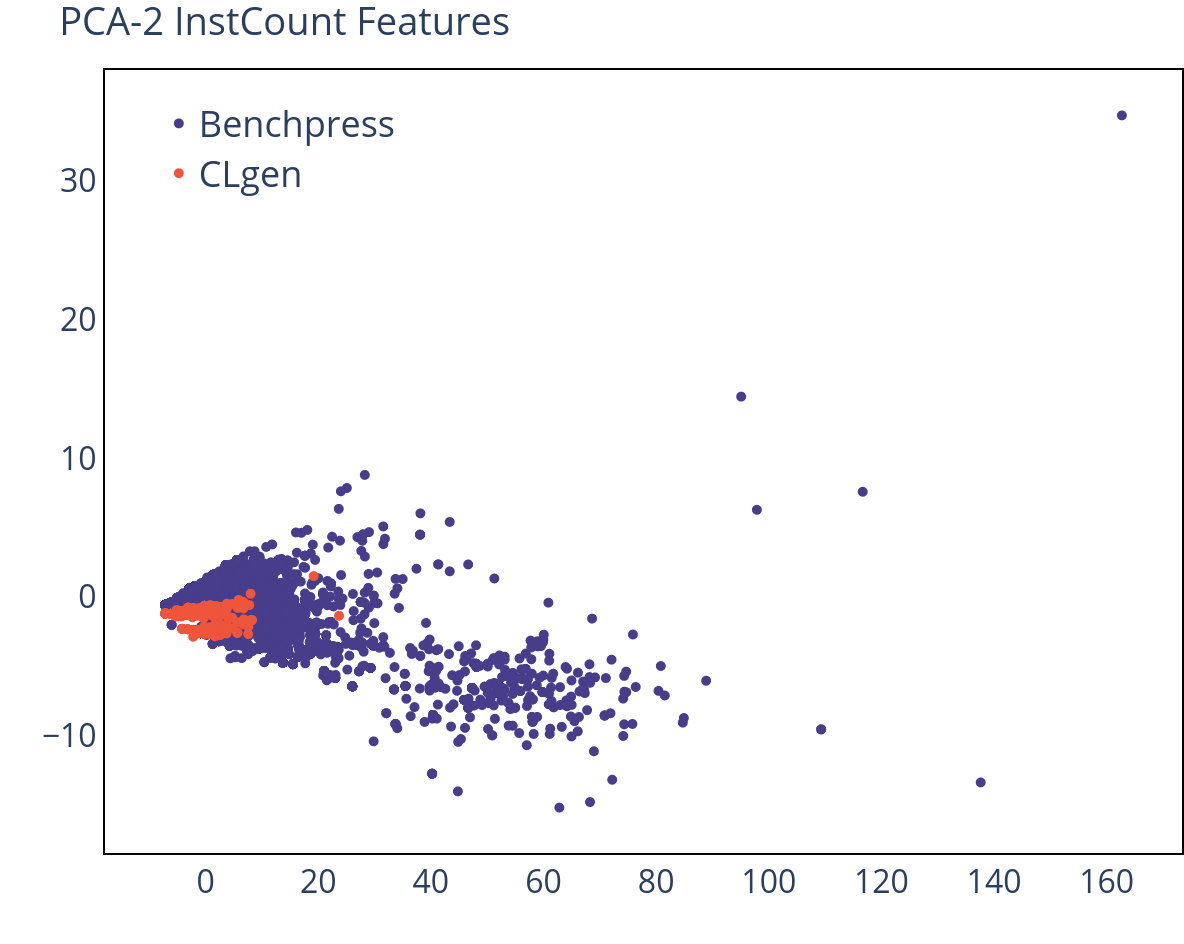}
        &
        \includegraphics[scale=0.139]{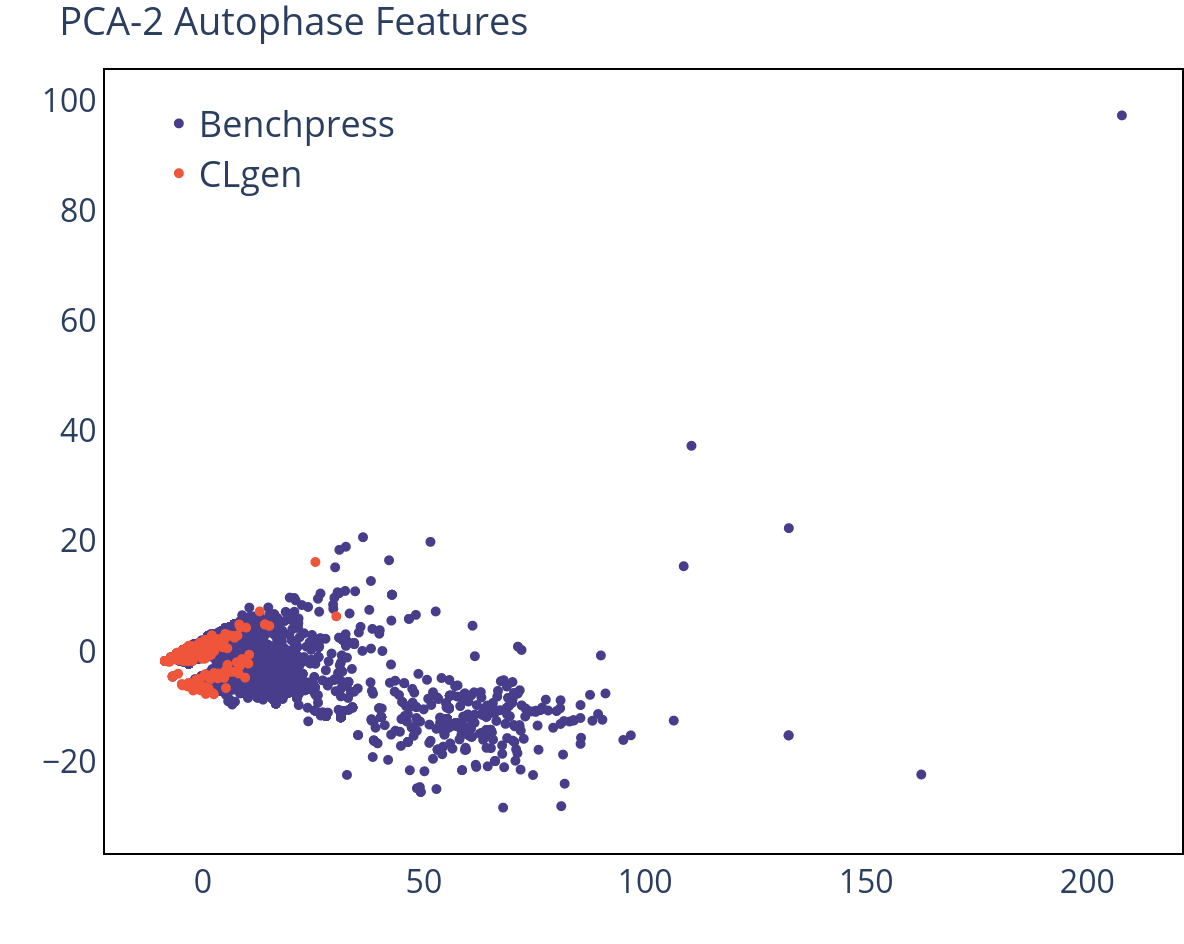}
        \\
	\end{tabular}  
	\caption{PCA-2 representation of feature space coverage of \name\ and \texttt{CLgen} for (a) Grewe's et al., (b) InstCount and (c) Autophase feature spaces. In this experiment, \name's generation is undirected and no iterative space search is performed.}
	\label{fig:pca_feature_coverage}
\end{figure*}

\paragraph{Compilation rate and code quality.}
\name\ generates over $10\times$ more unique compiling benchmarks than \texttt{CLgen}. This result is observed despite \name\ generating $8\times$ fewer unique benchmarks than \texttt{CLgen}. The compilation rate with \name\ is $86\%$ while \texttt{CLgen} has an exceedingly small rate of $2.3\%$.
\name's largest sample is 750 tokens compiling to 161 LLVM-IR instructions. This is a $7.5\times$ and $5\times$ increase in number of tokens and number of LLVM-IR instructions compared to \texttt{CLgen}'s largest kernel. The only drawback of \name\ compared to \texttt{CLgen} is that it is considerably slower in generating candidates. This is because the transformer-based architecture in \name\ is significantly larger in number of parameters than \texttt{CLgen}'s LSTM. Additionally, \name\ tends to generate longer kernels than \texttt{CLgen}, necessitating more inference steps and longer generation time.

In Figures~\ref{fig:size_distribution}a and~\ref{fig:size_distribution}b, we show the frequency distribution of the number of tokens and number of LLVM-IR instructions for compiling kernels for both datasets. To visualize our results better, we focus on synthesized kernels with token lengths $\leq 100$ and instructions lengths $\leq 25$ where the vast majority of benchmarks are found. Most of \name's benchmarks are found to have 20 to 80 tokens and 3 to 16 LLVM-IR instructions. The majority of \texttt{CLgen}'s benchmarks are found to have 5 to 45 tokens and only up to 4 LLVM-IR instructions. 94\% of \texttt{CLgen}'s generated benchmarks have only 1 instruction when compiled to LLVM-IR. We analyze the dataset to explain this phenomenon and find \texttt{CLgen} generates a lot of comments, repeated dead statements and awkward non-human-like code such as multiple semi-colons. These results agree with the case study by Goens et al.~\cite{goens} that shows the AST depth distribution of \texttt{CLgen}'s code is significantly narrower compared to code from \texttt{GitHub} or standard benchmarks.
\vspace{-5pt}
\paragraph{Feature space coverage.}
To further enhance our comparison, we perform an analysis on the feature space coverage of \name's and \texttt{CLgen}'s synthesized programs in all three feature spaces. Feature coverage is the most critical metric when evaluating the effectiveness of a benchmark synthesizer for predictive modeling. We use Principal Component Analysis (PCA-2) to represent the feature spaces in an easy to visualize 2-dimensional space. In Figures \ref{fig:pca_feature_coverage}a, \ref{fig:pca_feature_coverage}b and \ref{fig:pca_feature_coverage}c we show the extent of feature space covered by candidates in the two approaches. \texttt{CLgen}'s samples are clustered around the origin, while there is one outlier for Autophase and two for Grewe's et al. and InstCount features. Candidates generated by \name\ are more scattered achieving a much wider coverage of the feature space. 

\begin{figure*}[ht]
	\centering
	\setlength{\tabcolsep}{0.14em}
	\begin{tabular}{@{\vspace{-0.1em}}l@{}l@{\qquad}}
        \includegraphics[scale=0.195]{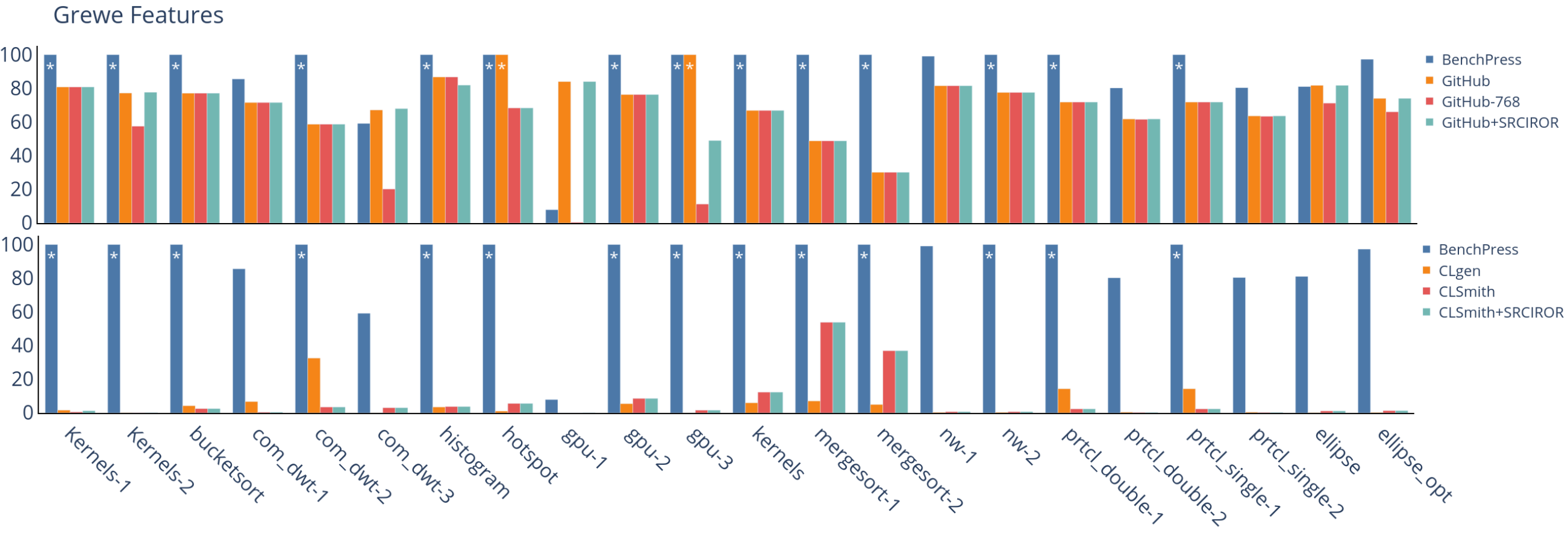}
        \\
        \includegraphics[scale=0.195]{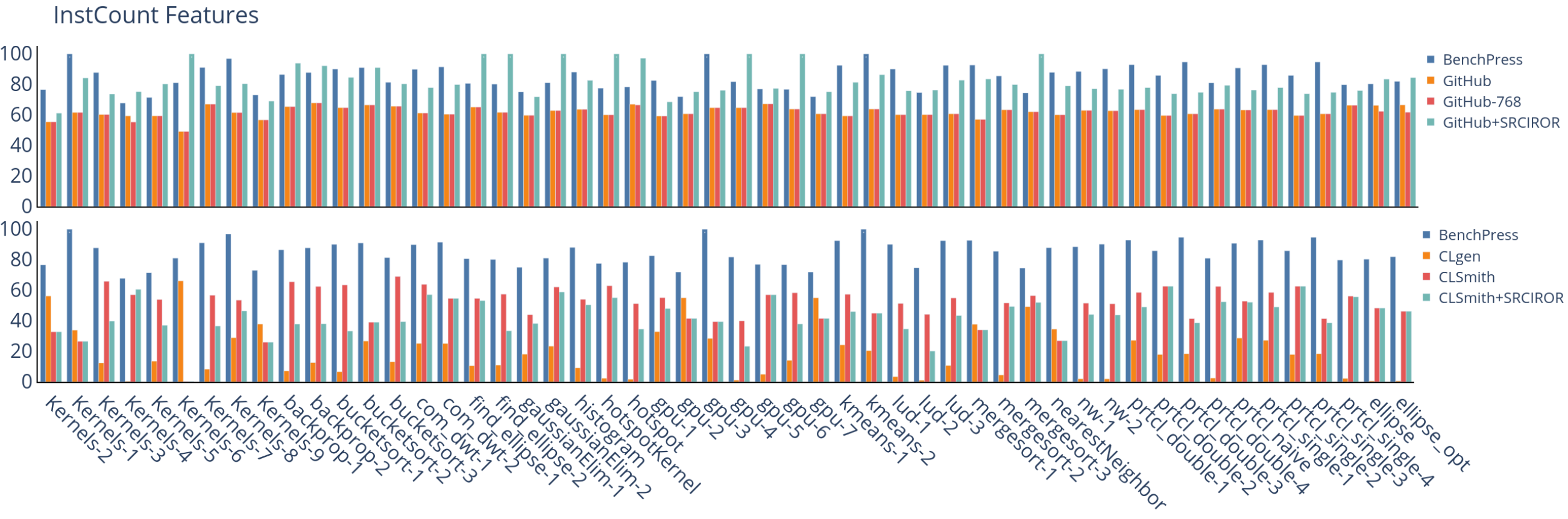}
        \\
        \includegraphics[scale=0.195]{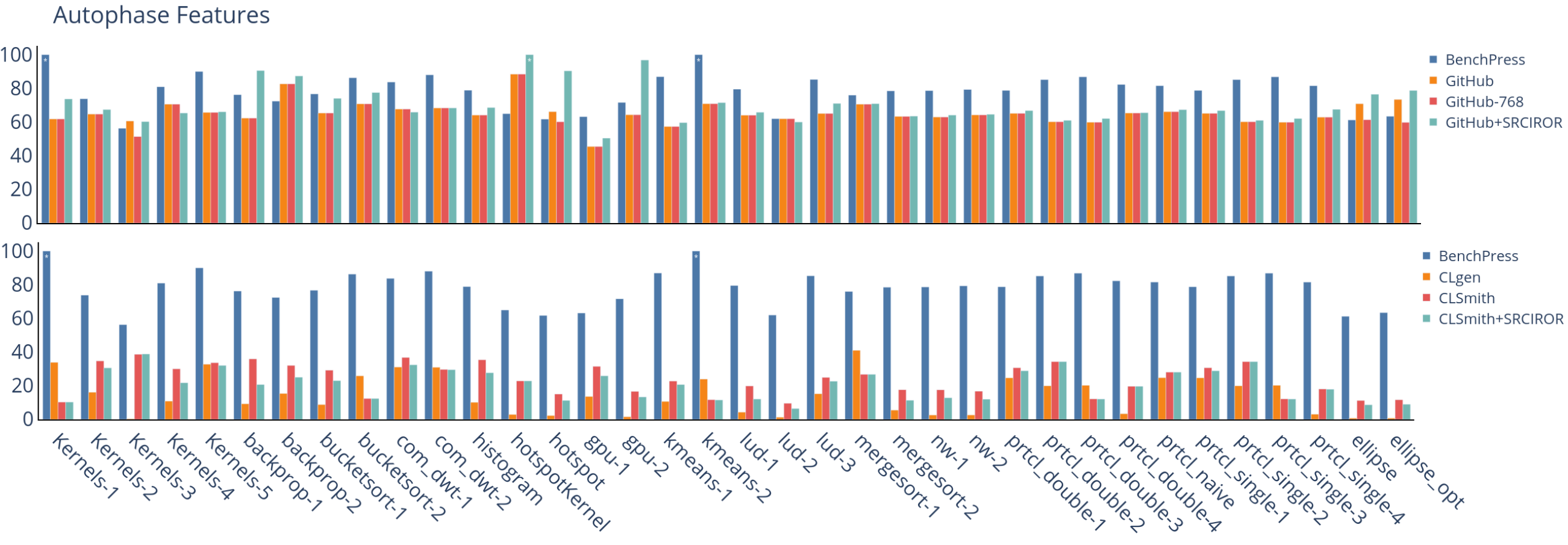}
        \\
	\end{tabular}
	\vspace{-10pt}
	\caption{Relative proximity to each Rodinia benchmark of the candidate kernel with the closest features. We report the best match for seven datasets (\name's, \texttt{CLgen}'s, \texttt{GitHub}'s and \texttt{GitHub-768}'s datasets also combined with exhaustive mutations with \texttt{SRCIROR})  over three feature spaces ((a) Grewe's et al., (b) InstCount and (c) Autophase). Relative proximity is 1 minus the distance of the two kernels in the feature space relative to the distance of the Rodinia benchmark from the axes origin. 
	100\% means an exact match in features and is highlighted with a white asterisk (*). A score towards 0\% indicates the closest match is closer to the axes origin than the benchmark, i.e., a very small or empty kernel.}
	\label{fig:search_results}
\end{figure*}

\subsection{Targeted Benchmark Generation}
\label{subsec:res_target}

We use beam search to generate samples that target desired parts of the feature space. We compare \name\ with human-written benchmarks from \texttt{GitHub} and synthetic benchmarks from \texttt{CLgen} and \texttt{CLSmith} in targeting the features of Rodinia benchmarks on three feature spaces. We use \texttt{SRCIROR} code mutator with beam search to collect \texttt{GitHub} and \texttt{CLSmith} benchmarks with closer features. For each target benchmark, we gather one OpenCL kernel per evaluated dataset whose features have the minimum available Euclidean distance from the target features. Figures \ref{fig:search_results}a, \ref{fig:search_results}b and \ref{fig:search_results}c show the relative proximity of each benchmark to the target. This proximity is the complement of the relative distance of the two kernels, i.e, 1 minus the distance between the two kernels in the feature space relative to the distance of the Rodinia kernel from the axes origin.
This allows us to express the quality of the match with an intuitive 0\% to 100\% scale: 100\% means the two kernels have the same features, 0\% means the best kernel is as close to the target as an empty kernel. We mark perfect matches with a white asterisk (*).
\vspace{-5pt}
\paragraph{Performance on syntactic features.} On Grewe's et al. feature space, \name\ generates kernels that are the closest ones in features for all 22 Rodinia Benchmarks compared to \texttt{CLgen} and \texttt{CLSmith}, and 20 out of 22 compared to \texttt{GitHub} and \texttt{GitHub-768}. \name\ synthesizes an exact match (100\% relative proximity) for 14 target benchmarks.

We pick out and discuss a few examples from our results. The absolute distance achieved for `nw-1' and `ellipse\_opt', is $1.0$. For both targets, almost all features match except for one missing instruction (\texttt{coalesced mem access} and \texttt{atomic inst} respectively). For `hotspot' \texttt{GitHub} and \name\ produce a candidate kernel with exact matching features. However, \name\ generates the matching candidate kernel in 421 tokens, unlike \texttt{GitHub}'s closest benchmark that has 798 tokens. For the two target benchmarks that \name's candidates were not closest to, we found only \texttt{GitHub} contains better samples for `com\_dwt-3' and \sloppy{and `gpu-1'}, while \name\ does not. We find both benchmarks to be fairly large (901 and 5,200 tokens respectively) and \name\ cannot reach these features within 768 tokens. For the same reason, \texttt{GitHub-768}, \texttt{CLgen} and \texttt{CLSmith} does worse than \name\ on these targets.

\begin{figure}[ht]
\centering
    \includegraphics[scale=0.182]{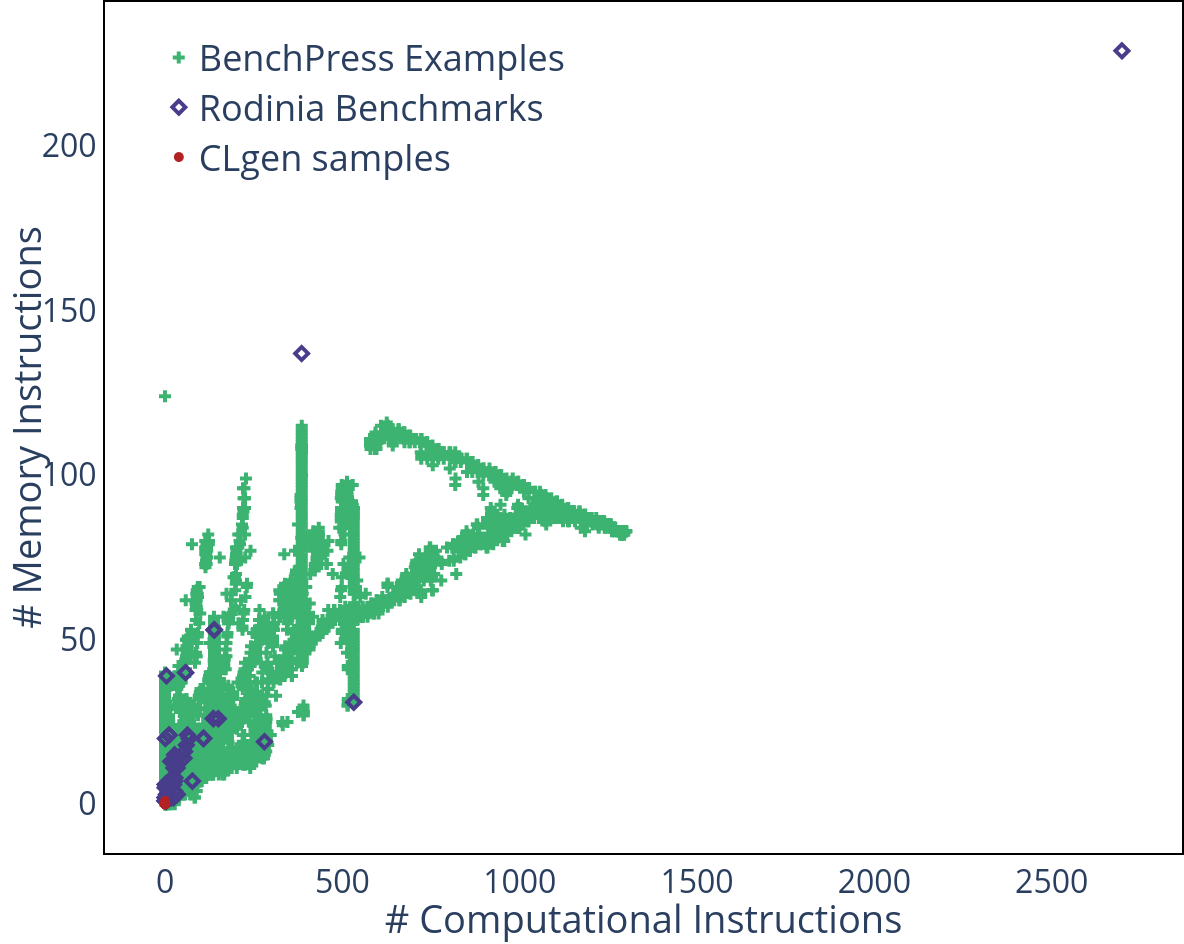}
    \caption{\# Memory operations and \# computational instructions for \textbf{(a)} Rodinia benchmarks in purple diamonds, \textbf{(b)} \texttt{CLgen}'s samples in red dots and \name's benchmarks in green crosses after performing directed search for all Rodinia benchmarks.}
    \label{fig:coverage_mem_comp}
\end{figure}

\paragraph{Performance on LLVM IR features.} Autophase and InstCount features are extracted from the LLVM-IR of a program that has been compiled with \texttt{-O1} flag to apply basic optimisations such as dead code elimination. \name\ occasionally generates repeating operations that a compiler will remove or numerical operations that may be reduced to simple assignments. Owing to these optimisations, we find targeting benchmarks on these two feature spaces is more challenging than Grewe's et al. syntax-level features. With InstCount features, \name\ generates candidates whose features completely match 2 out of the 52 Rodinia benchmarks. Among the remaining 50, \name\ outperforms \texttt{CLgen}, \texttt{CLSmith}, \texttt{GitHub} and \texttt{GitHub-768} for all target benchmarks, achieving higher proximity. \texttt{SRCIROR} significantly improves \texttt{GitHub} leading to \texttt{GitHub+SRCIROR} to achieve better proximity for 18 out of 52 Rodinia benchmarks compared to \name. On Autophase features, \name\ generates candidates matching the same 2 target benchmarks, while outperforming \texttt{CLgen}, \texttt{CLSmith} and \texttt{GitHub} on 30 out of 36 Rodinia benchmarks in total. \texttt{GitHub+SRCIROR} performs better than \name\ for 8 out of 36 target benchmarks and produces an exact match for `hotspotKernel'.

We previously explain the importance of having diverse features in compiler benchmarks and we show, in Figure \ref{fig:motivating_example}, how sparse Rodinia benchmarks are on Grewe's et al. reduced feature space and how \texttt{CLgen} fails to provide any additional features. Now we introduce into this 2-dimensional space all \name's kernels that are generated while performing directed space search to target Rodinia benchmarks and we present them in Figure~\ref{fig:coverage_mem_comp}. \name\ densely populates the space around the target benchmarks that are clustered around the lower left corner. We find \name's samples progressively converge to the target benchmark features with successive generations. 
For example, \name\ targets `com\_dwt-3' at 385 computational and 137 memory instructions, starting from the axes origin and attempting to reach its features from different directions. One of the directions prevail but does not manage to exactly reach the target. The same happens for the top right point, `gpu-1'. \name's samples get closer developing a straight line from the origin to 1,000 computational and 100 memory instructions. At this point \name\ is restricted by its sequence length and cannot augment further its samples. This is depicted by its attempt to reduce the distance by swapping the two instruction types within the same token length, forming a perpendicular line with a negative slope. We argue the area of Grewe's et al. feature space that \name\ can cover within 768 tokens to be the area of the triangle formed by the intersections of the axes with the extension of the negative slope line developed by \name's samples.

\paragraph{Summary - \name\ vs GitHub vs CLgen vs CLSmith.} 6 of the targeted Rodinia benchmarks exceed \name's maximum sequence length of 768 tokens. In LLVM-IR feature spaces, care must be taken to generate code that will not be removed by compiler optimisations. This is a difficult challenge for source code generative models. However, our results demonstrate that \name\ can generate OpenCL kernels that approach target human-written benchmarks compared to \texttt{GitHub} code and CLgen candidates. Our experiments also show \name\ is dramatically better in all cases than \texttt{CLgen}, the current state of the art in OpenCL synthetic benchmark generation. We further elaborate on \name's performance in the next subsections.

\subsection{Active Learning for Feature Selection}
We combine \name's ability to generate benchmarks targeting desired features with active learning in order to generate benchmarks that improve the training of the Grewe et al. heuristic. We evaluate this against passive training with \texttt{CLgen}, \texttt{GitHub} code, and \name\ with randomly selected target features. All approaches augment the same baseline training set that is taken from~\cite{clgen}, containing 7 benchmark suites\footnote{The benchmarks have been updated with a wider range of \texttt{global} and \texttt{local sizes}.}. 
Table~\ref{tab:grewe_heuristic} shows the effect of each approach on the predictive power of the heuristic. Training only on human written benchmarks improves the heuristic's performance by 4\%, as shown in Table~\ref{tab:grewe_heuristic}'s first row. To understand the maximum achievable improvement in the heuristic, we compute the best speedup ($=12\%$) that is achieved if the model chooses the optimal device as opposed to always picking the GPU. For 71\% of the benchmarks, GPU is the optimal device, so no speedup improvement is possible. For the remaining 29\% benchmarks, predicting the `CPU' label correctly with Grewe et al. will result in a speedup improvement. 

\begin{table}[]
\begin{tabular}{lllll}
           & \begin{tabular}[c]{@{}l@{}}Speedup \%\end{tabular} & \begin{tabular}[c]{@{}l@{}}Precision\end{tabular} & \begin{tabular}[c]{@{}l@{}}Recall\end{tabular} & \begin{tabular}[c]{@{}l@{}}Specificity\end{tabular}  \\\hline
\texttt{Benchmarks} & +4\% & 0.81 & 0.86 & 0.61 \\
\texttt{BenchPress-AL} &  +6\% & 0.84 & 0.86 & 0.64 \\
\texttt{BenchPress-P} &  +1\% & 0.84 & 0.85 & 0.48 \\
\texttt{CLgen}      & -1\% & 0.52 & 0.86 & 0.43 \\
\texttt{GitHub}     & +1\% & 0.85 & 0.83 & 0.61 
\end{tabular}
\caption{Grewe et al. heuristic model's performance, precision, recall, and specificity when trained on each technique. Speedup is the geometrical mean of speedups over all benchmarks relative to the optimal static decision, i.e. running on the GPU. Precision, recall, and specificity treat GPU labels as positive and CPU labels as negative.}
\label{tab:grewe_heuristic}
\end{table}

\begin{figure}[ht]
\centering
    \includegraphics[scale=0.21]{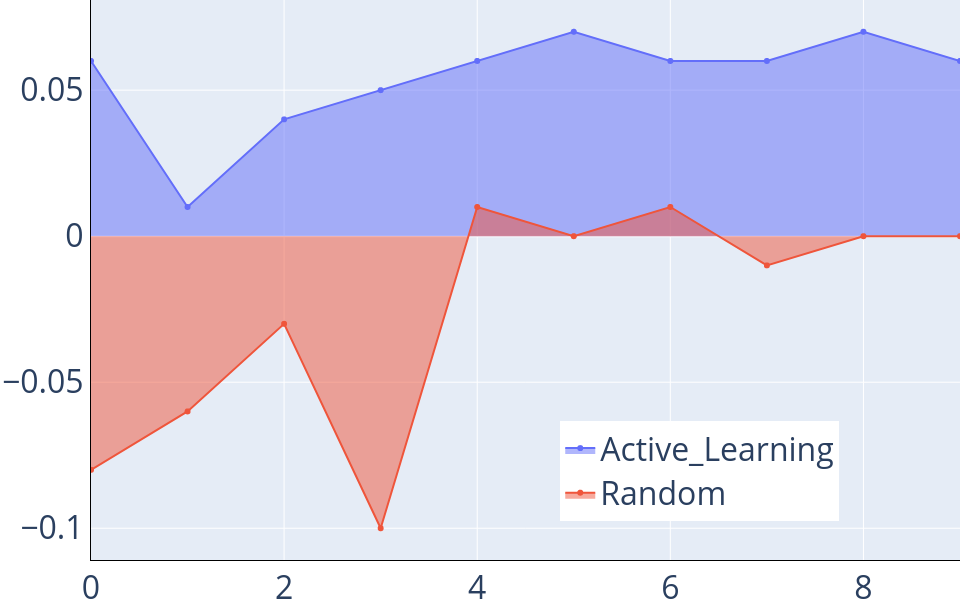}
    \caption{\name's performance enhancement of Grewe et al. heuristic model when using active learning compared to passively targeting random parts of the feature space over the course of 10 sampling epochs.}
    \label{fig:passive_vs_active}
\end{figure}

\name\ using active learning (\texttt{BenchPress-AL}) clearly outperforms all other approaches in terms of average speedup, improving it by 6\%. When trained on \texttt{BenchPress} with passive/random feature selection (\texttt{BenchPress-P}), the speedup achieved is only 1\%. To our surprise, the same speedup is achieved with \texttt{GitHub}, which is worse compared with training only on the original benchmark suites. We further analyze the dataset collected from \texttt{GitHub} code and we find it to be imbalanced with 90\% of its training instances are labelled as `GPU'. This leads the model having a higher precision of 0.85, i.e. predicting correctly that a kernel should execute on the GPU, but falling short when it comes to correctly predicting the `CPU' label. Training the heuristic with \texttt{CLgen} actually leads to a slowdown: it is 1\% slower to execute kernels on the predicted devices compared to statically executing everything on the GPU, the baseline device. We analyze \texttt{CLgen}'s dataset and observe the opposite pattern found in \texttt{GitHub}'s dataset. 63\% of its training data execute faster on the CPU than on the GPU. This is a direct consequence of \texttt{CLgen} generating small benchmarks that are poor in features, as the CPU may be slower than the GPU but the large overhead of transferring data to the GPU makes the CPU a better choice for small workloads. \texttt{CLgen} containing too many CPU-labeled kernel explains the heuristic's low precision and specificity, as it becomes biased to select the CPU very often leading to a slowdown.

\begin{figure*}[ht]
    \centering
    \begin{tabular}{cc}
        \includegraphics[scale=0.38]{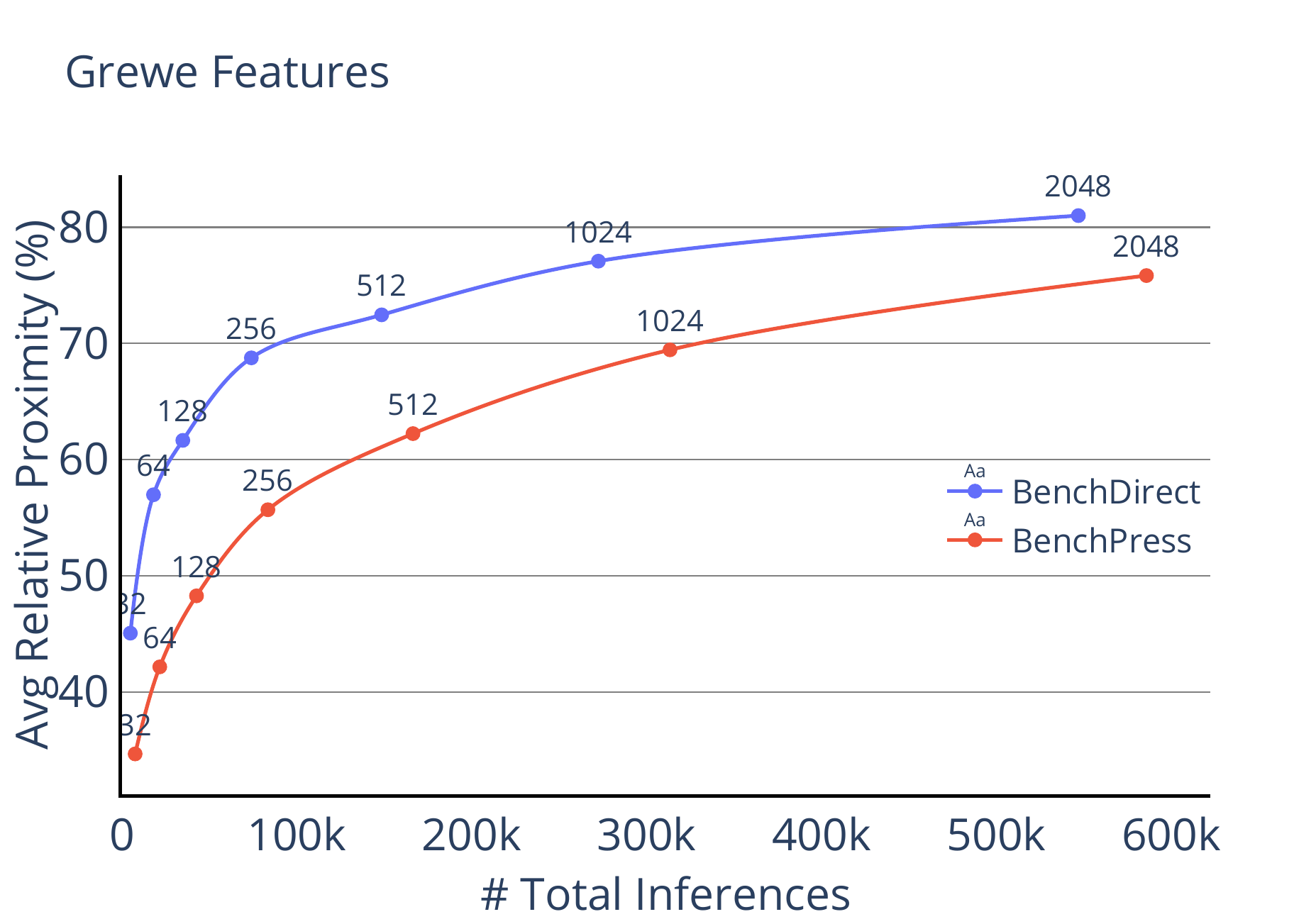}
        &
        \includegraphics[scale=0.38]{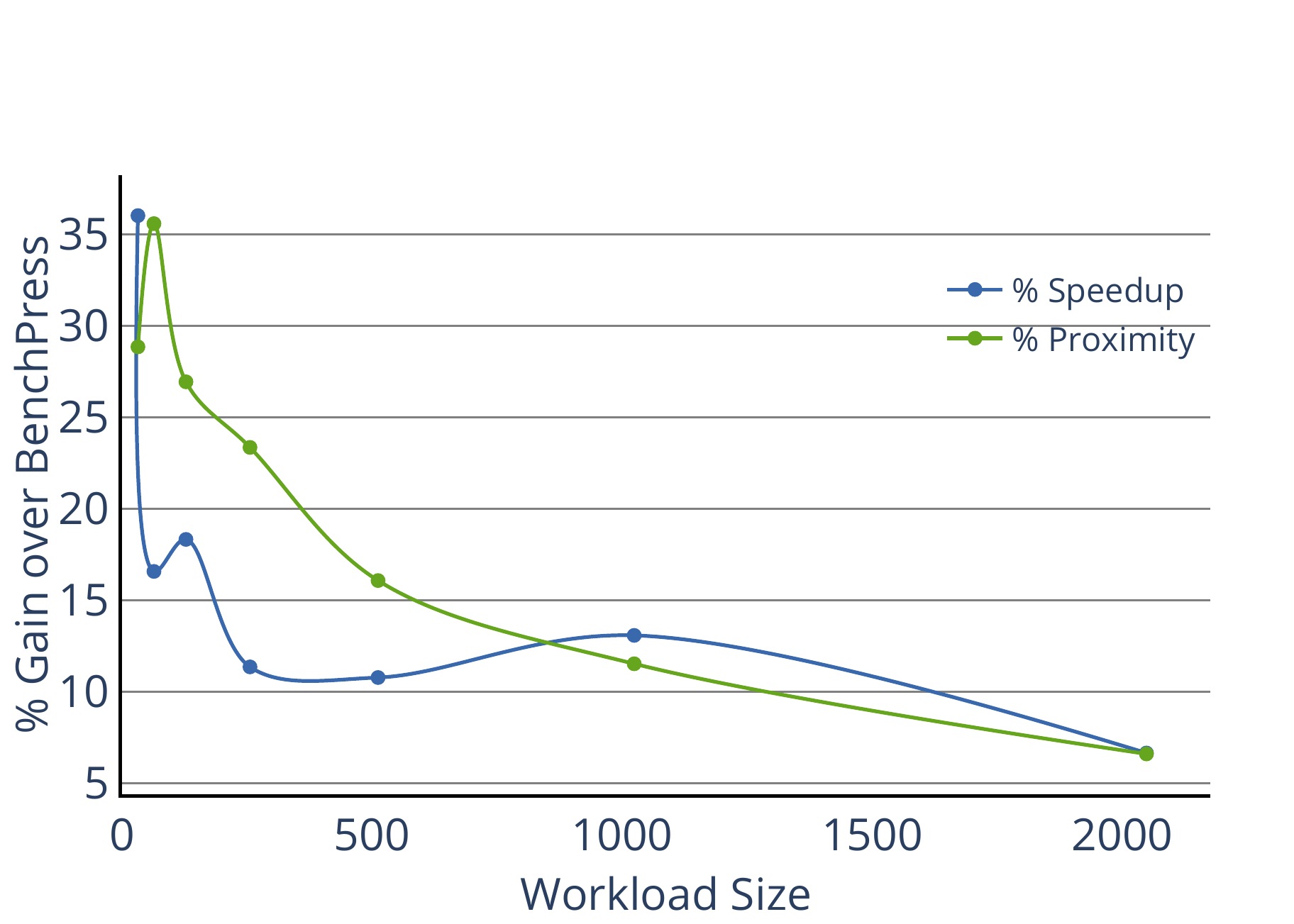}
        \\
        \includegraphics[scale=0.38]{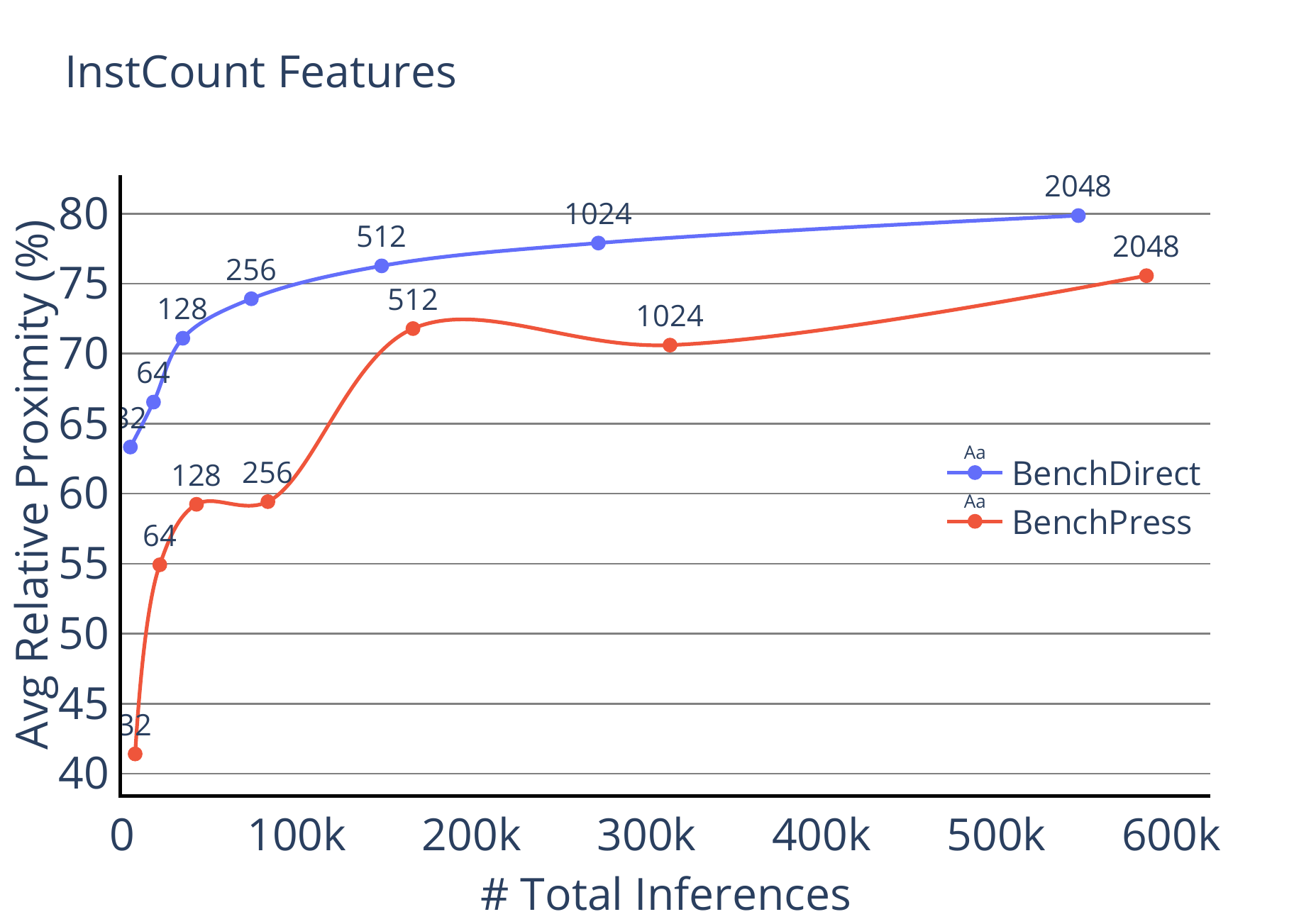}
        &
        \includegraphics[scale=0.38]{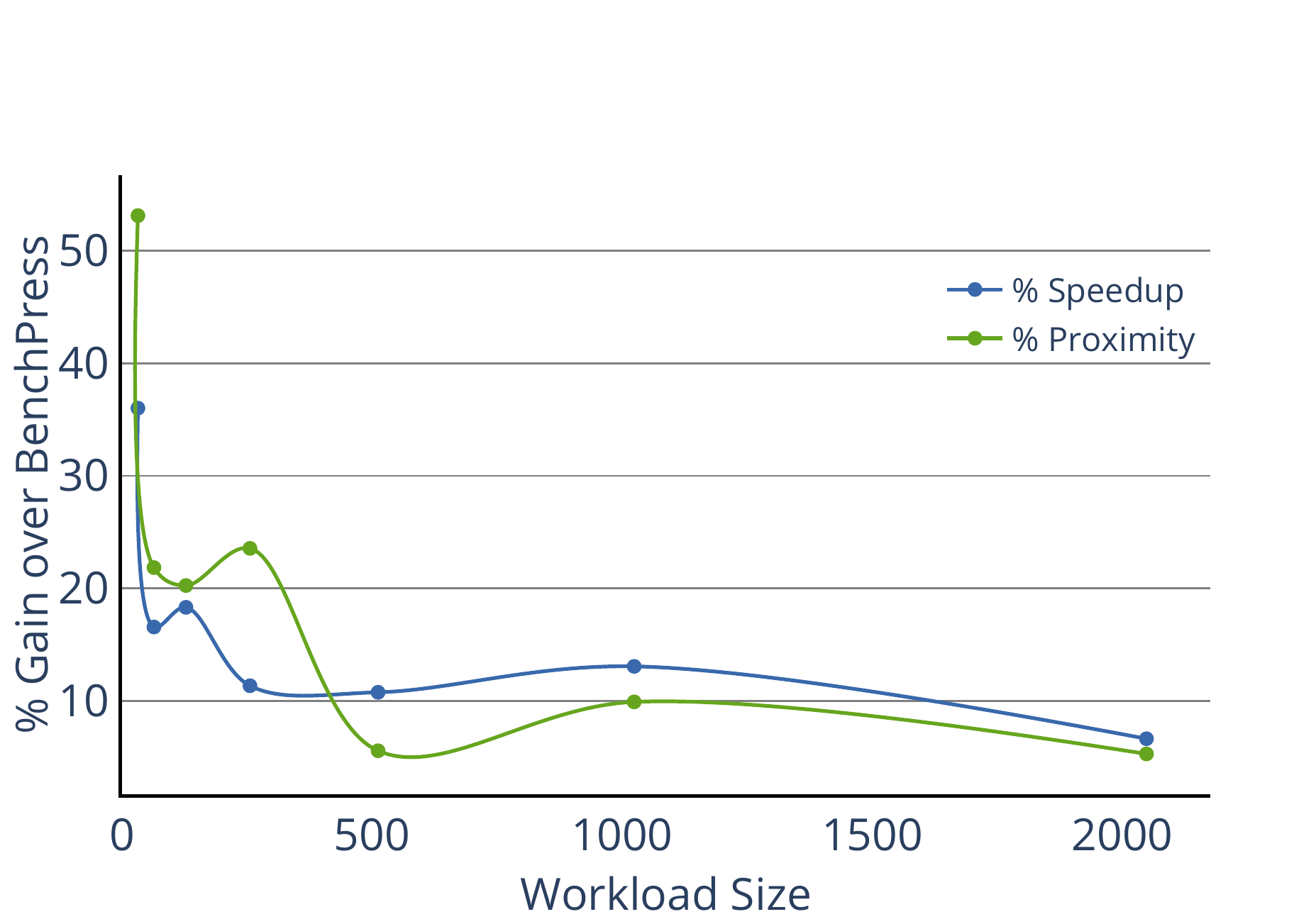}
        \\
        \includegraphics[scale=0.38]{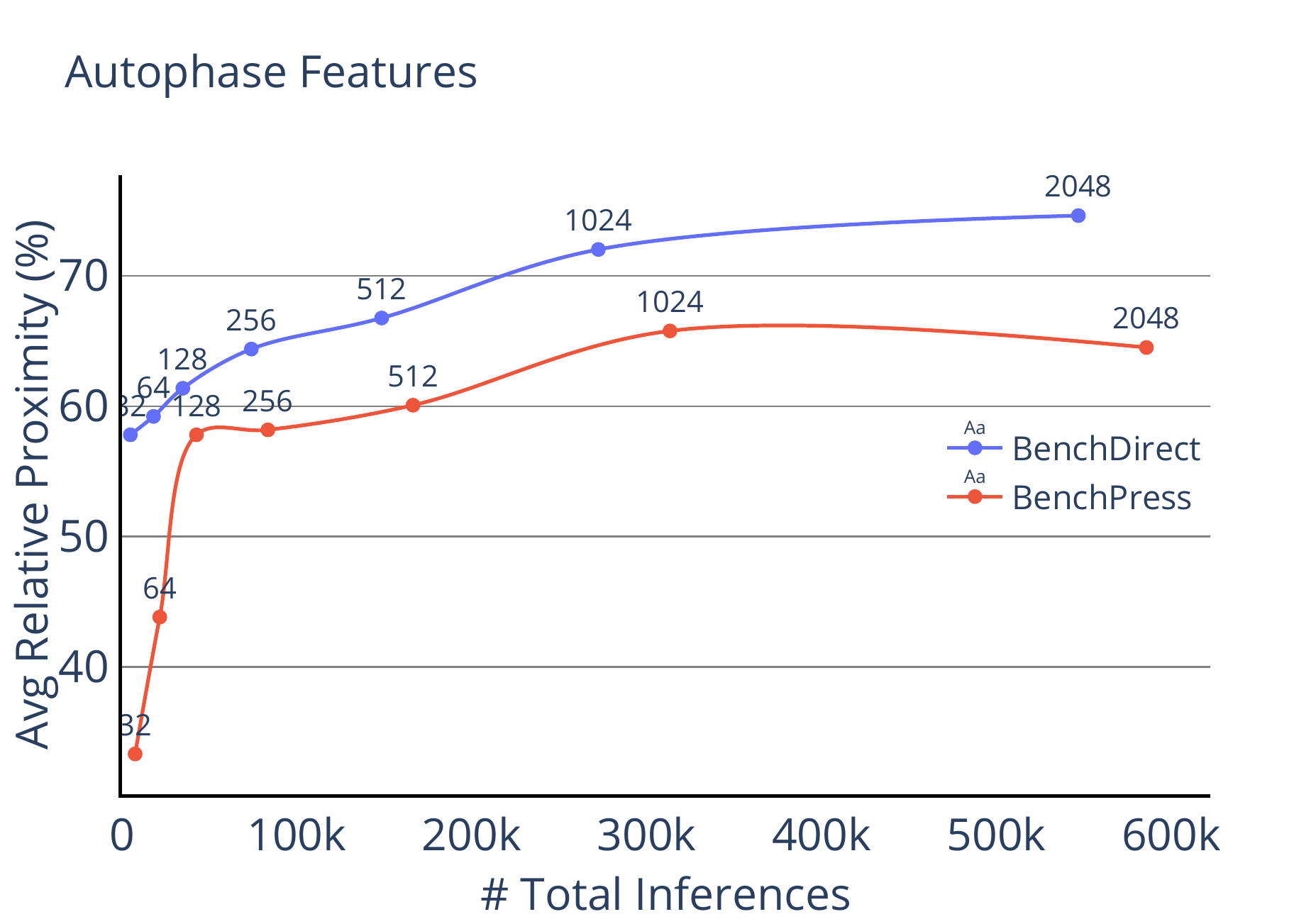}
        &
        \includegraphics[scale=0.38]{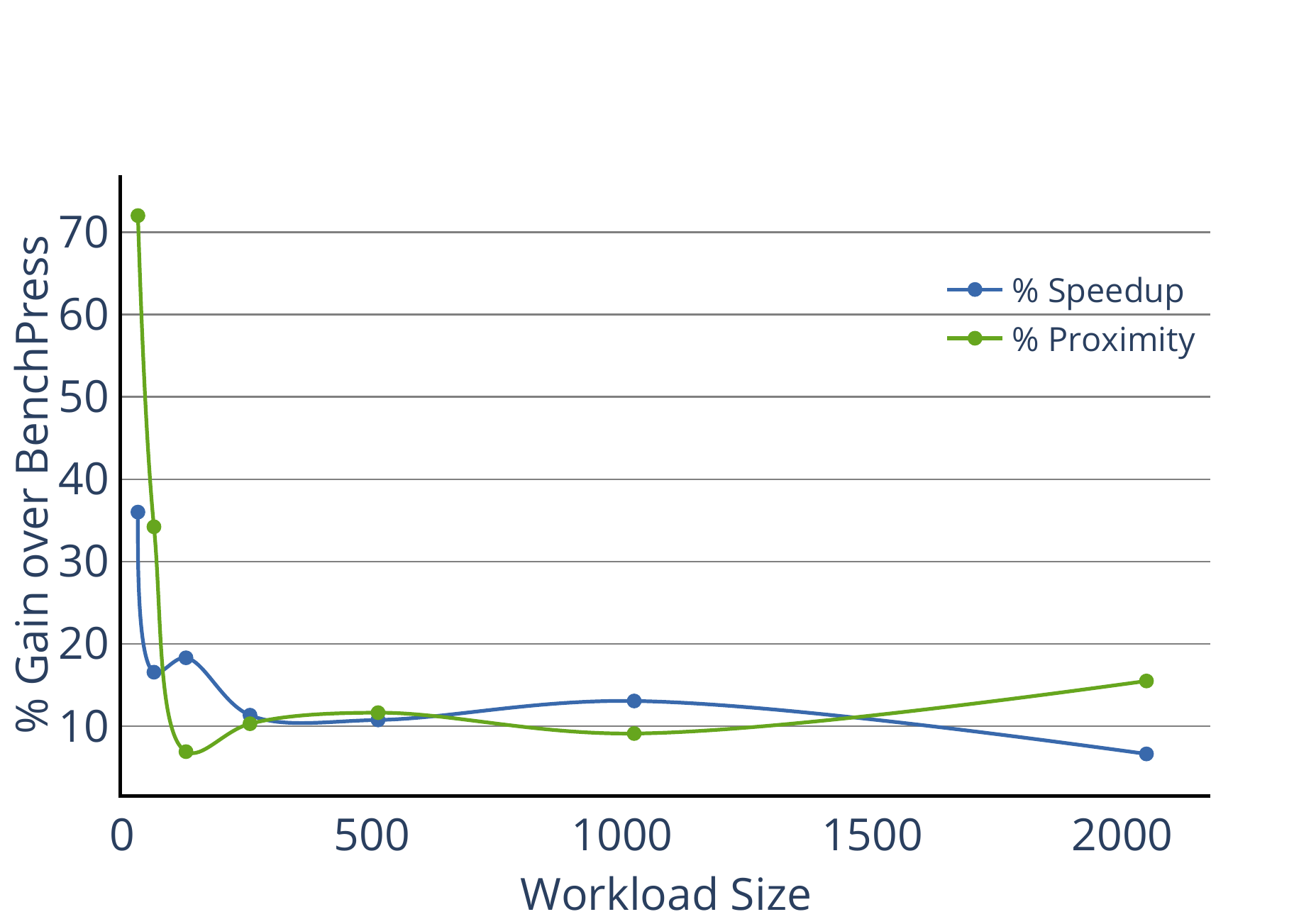}
        \\
    \end{tabular}  
    \vspace{-10pt}
    \caption{Pareto fronts of the average relative proximity versus total inferences in targeting Rodinia benchmarks over three feature spaces ((a) Grewe's et al., (b) InstCount and (c) Autophase). Higher relative proximity and fewer inferences are better, therefore optimal points, i.e., Pareto-dominant, are those towards the top left. We annotate the workload size configuration per Pareto point. On the right, we show \texttt{BenchDirect}'s acquired speedup and accuracy gain over \name\ for the same workload size setting.}
    \vspace{-5pt}
    \label{fig:pareto}
\end{figure*}

Our main motivation behind using active learning is that it gives \name\ the ability to target directly those parts of the feature space that will maximize a downstream task's performance. To assess the active learner's performance, we compare the Grewe et al. heuristic's speedup when trained on \name's benchmarks that target areas of the feature space selected by the active learner versus benchmarks that target random features. In both cases, we execute \name\ for the same amount of time, 10 sampling epochs (i.e., performing steered generation for 10 target feature vectors). In Figure~\ref{fig:passive_vs_active}, we show the speedup achieved by the heuristic when trained on the data collected at that step. Using active learning to target features, \name's dataset improves the heuristic's speedup by 50\% after 5 sampling steps, from 4\% to 6\%. Targeting random features never leads to a speedup higher than 1\%. \name\ can still develop the same speedup by targeting random features if infinite amount of time was available. Our active learner ensures that missing features are going to be quickly targeted, improving the state of the art within 5 sampling epochs.

\subsection{Directed Language Modeling}
\label{subsec:res_target_direct}

\begin{figure*}[!htbp]
	\centering
	\setlength{\tabcolsep}{0.14em}
	\begin{tabular}{@{\vspace{-0.6em}}l@{}l@{\qquad}}
        \includegraphics[scale=0.275]{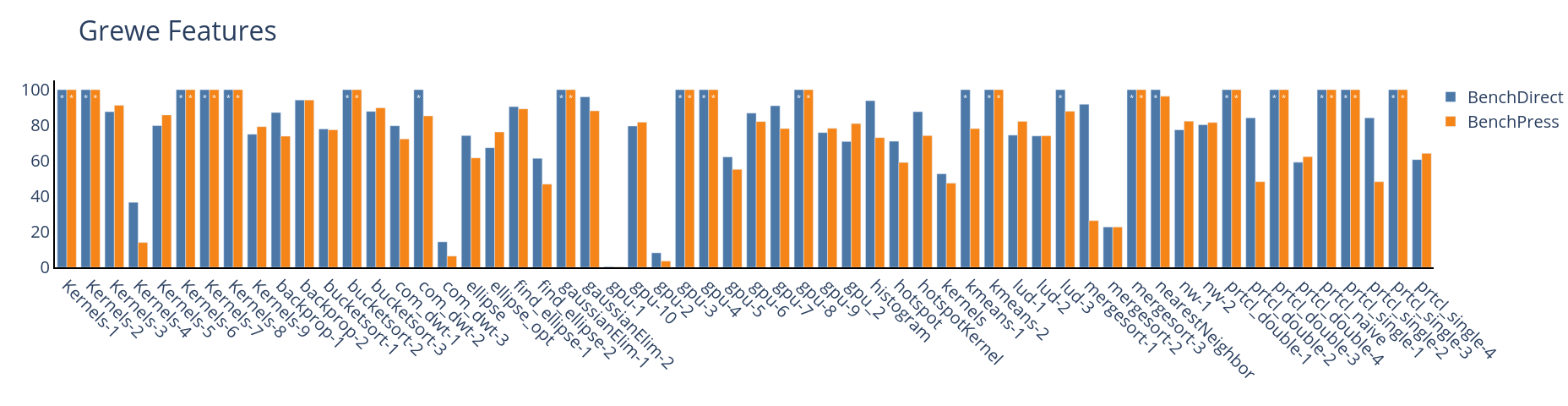}
        \\
        \includegraphics[scale=0.275]{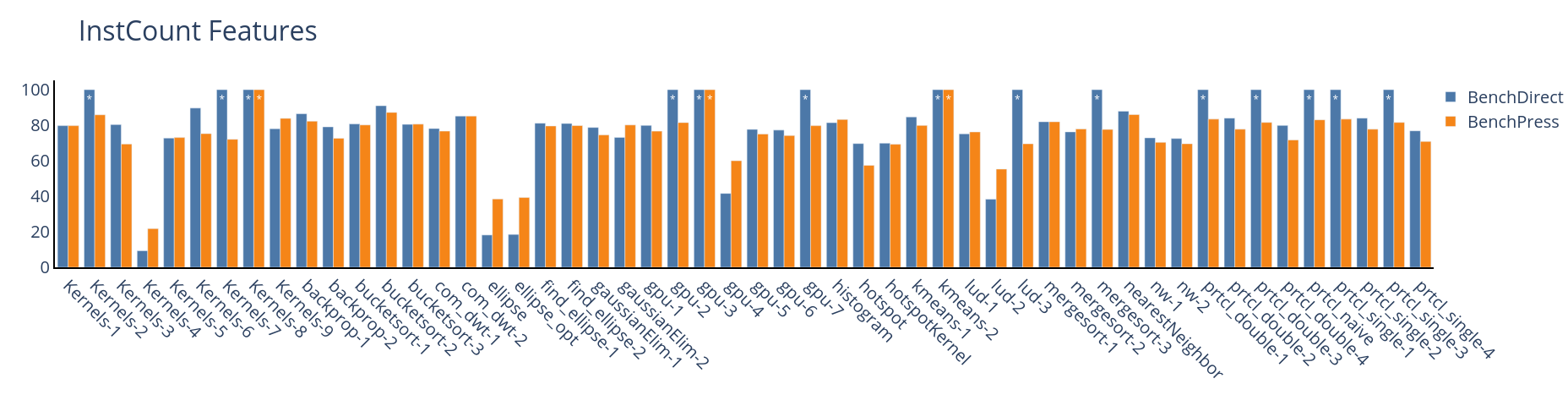}
        \\
        \includegraphics[scale=0.275]{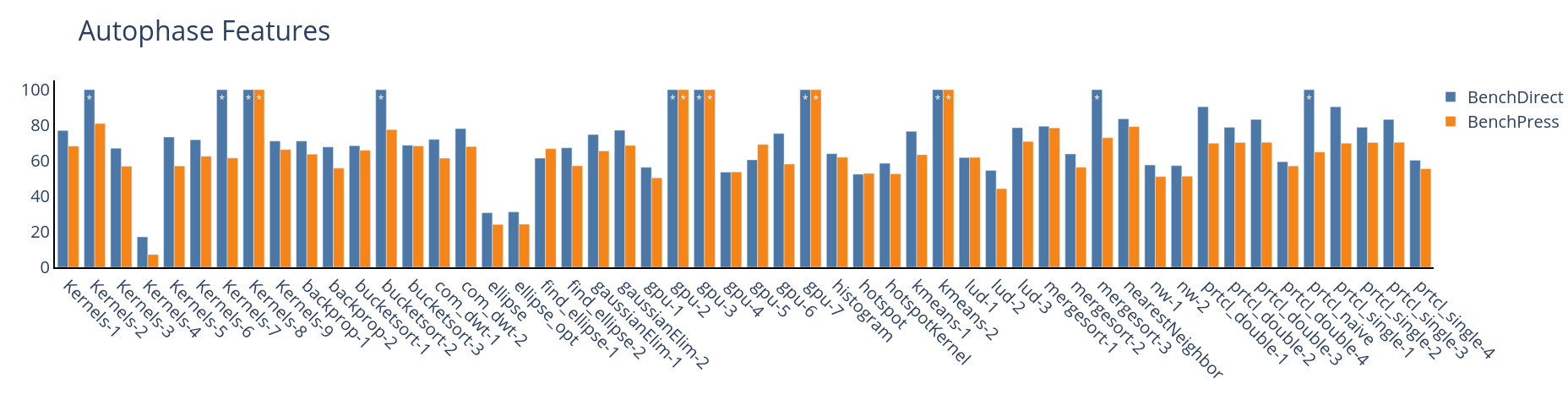}
        \\
	\end{tabular}
	\vspace{-10pt}
	\caption{Relative proximity to each Rodinia benchmark of the candidate kernel with the closest features. We show the best match for \texttt{BenchDirect} and \name. Relative proximity is defined in figure~\ref{fig:search_results}.}
	\label{fig:directed_search_results}
\end{figure*}

We target the features of Rodinia benchmarks using \name\ and \texttt{BenchDirect}. Both models use beam search over their synthesizer to minimize their samples' distance from the target features. At the end of each search, we select the generated kernel whose features have the minimum Euclidean distance from the target benchmark. We perform this experiment for multiple beam search candidate sizes: 32, 64, 128, 256, 512, 1024 and 2048. On the left side of Figures~\ref{fig:pareto}a, ~\ref{fig:pareto}b and ~\ref{fig:pareto}c we show the Pareto fronts of the average relative proximity achieved over all Rodinia benchmarks versus the total amount of inferences. Relative proximity is defined in Section~\ref{subsec:res_target} as a percentage of how close a feature vector is to the target features relatively to the axis origins. Inferences are calculated as the number of beam search iterations to target all benchmarks multiplied by the workload size. Each datapoint is annotated with its workload size configuration. On the right side of Figures~\ref{fig:pareto}a, ~\ref{fig:pareto}b and ~\ref{fig:pareto}c, we show \texttt{BenchDirect}'s improvement in accuracy and execution time compared to \name, for each workload size setting.

\texttt{BenchDirect} outperforms \name\ in average relative proximity and total inferences for all workload size configurations, across all three feature spaces. Taking the average proximity and the execution time as a design space, the datapoints that are optimal with respect to these two metrics belong exclusively to \texttt{BenchDirect}, while there are no configurations for \name\ that optimise either metric compared to the former. The effect \texttt{BenchDirect}'s directed language model has in targeting features is especially denoted when the workload size is small. \texttt{BenchDirect}'s synthesizer conditions directly on the target features and provides, in very few attempts, candidates that match or are very close to them. This means a dramatic reduction in the amount of benchmarks per beam search does not drastically hamper the model's accuracy. The same is not true for \name. While \texttt{BenchDirect} offers an average speedup of 10.2\% and an improvement in average relative proximity of 10.1\% for workloads greater or equal to 512, the speedup reaches up to 36\% in all three feature spaces and the accuracy gain up to 72\% on InstCount features for smaller workloads. This indicates \texttt{BenchDirect} remains consistent in the amount of iterations needed to achieve high accuracy, while \name\ suffers in both areas.

\begin{figure*}[!htbp]
    \centering
    \begin{tabular}{ccc}
        \includegraphics[scale=0.12]{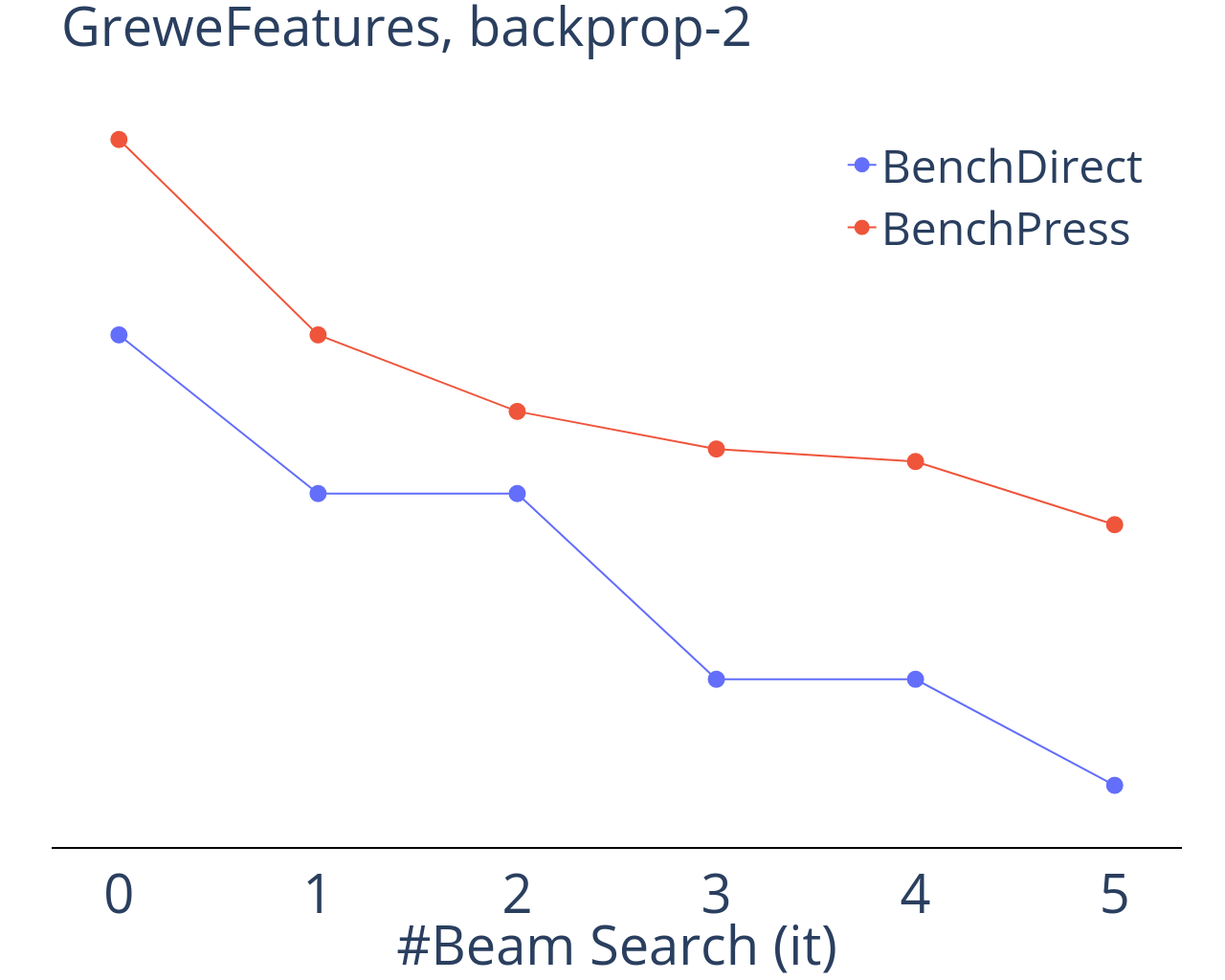}
        &
        \includegraphics[scale=0.12]{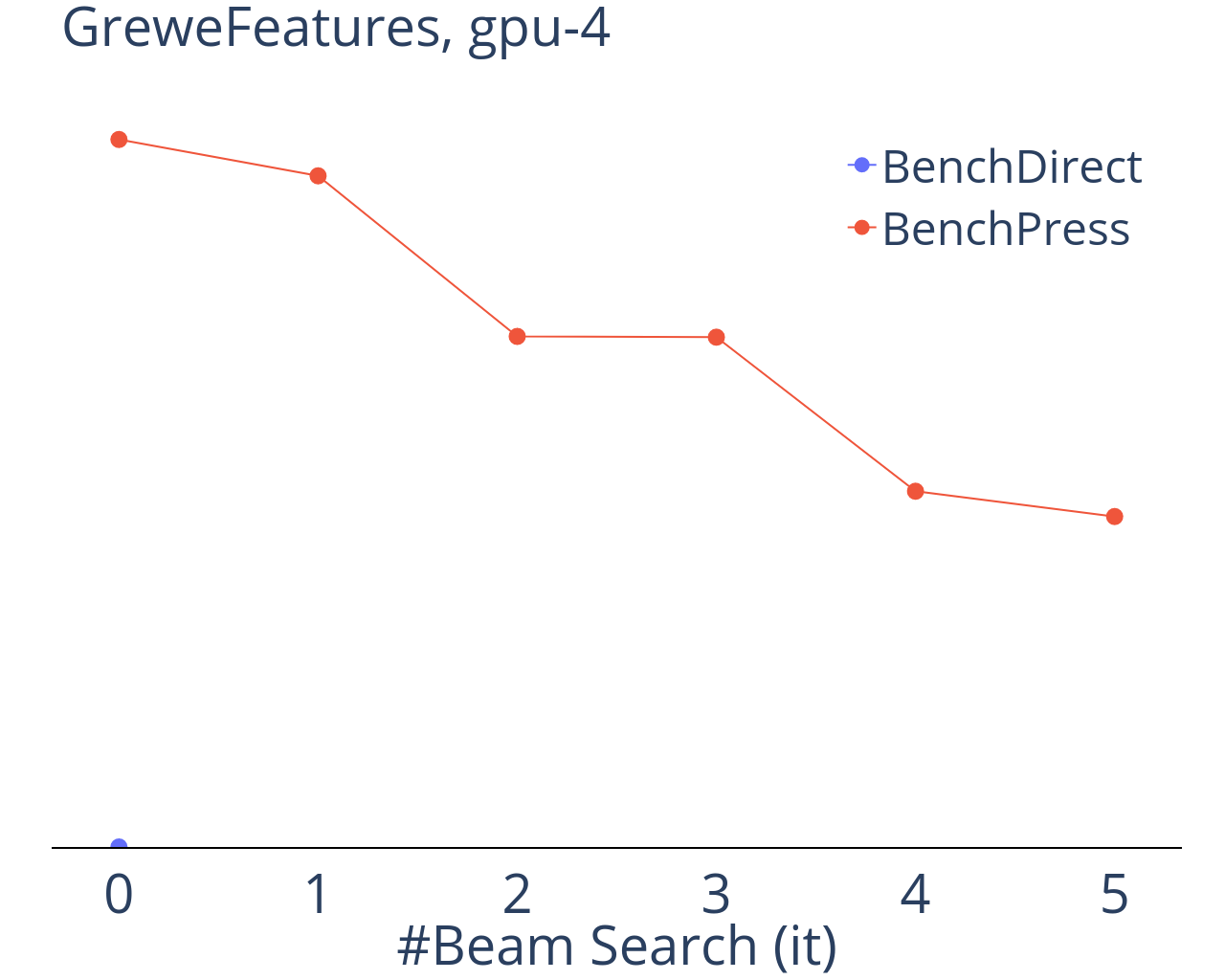}
        &
        \includegraphics[scale=0.12]{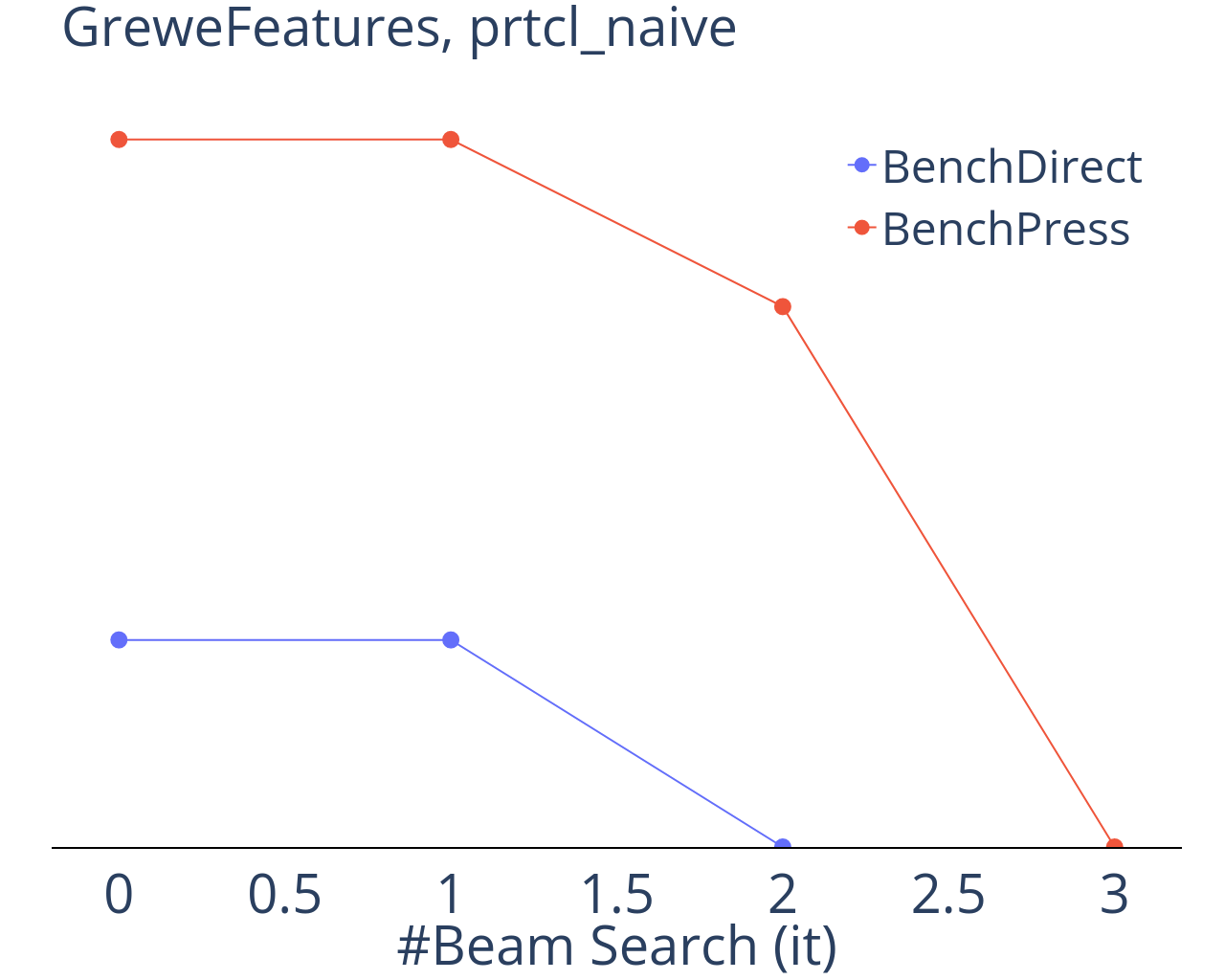}
        \\ [0.5cm]
        \includegraphics[scale=0.12]{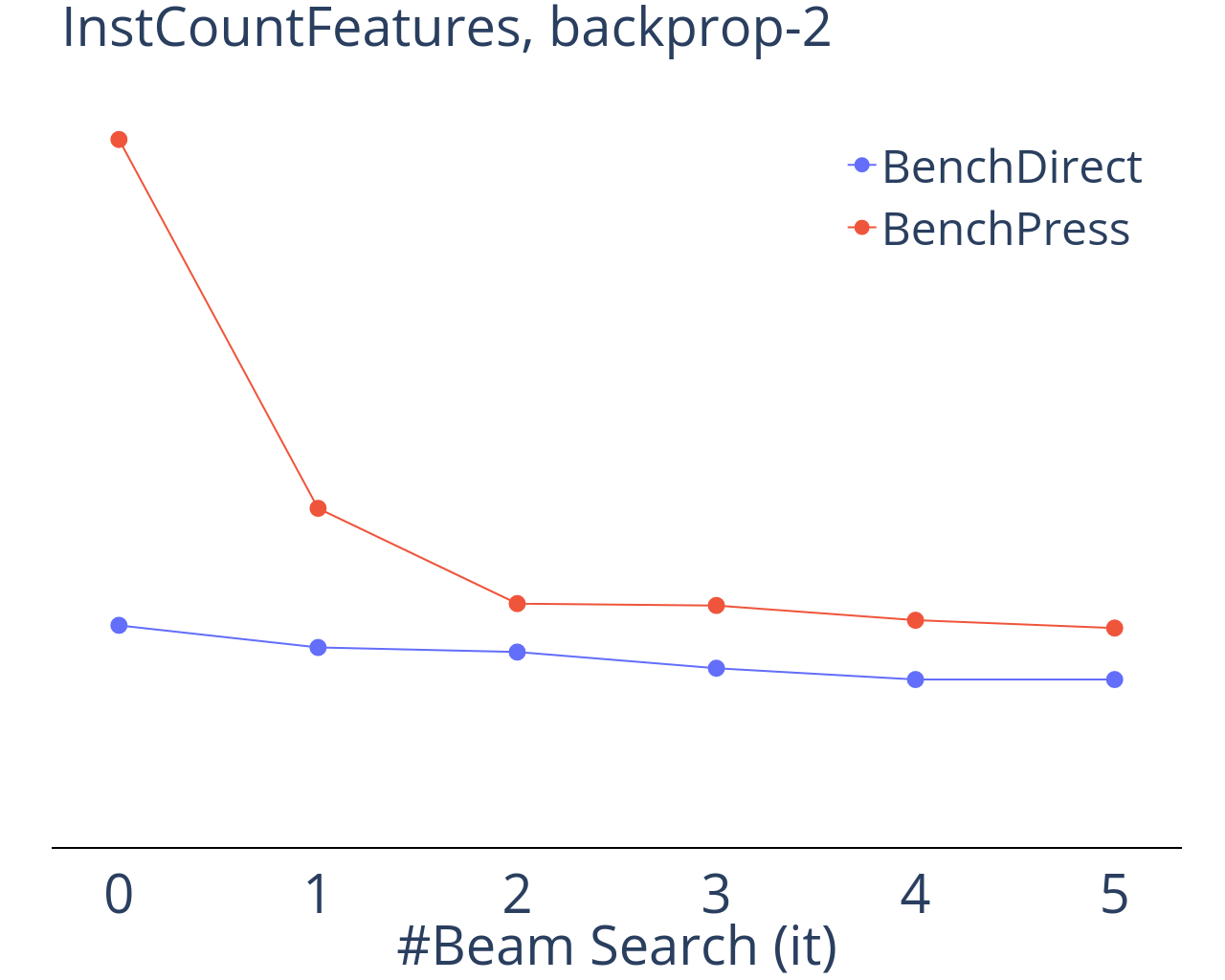}
        &
        \includegraphics[scale=0.12]{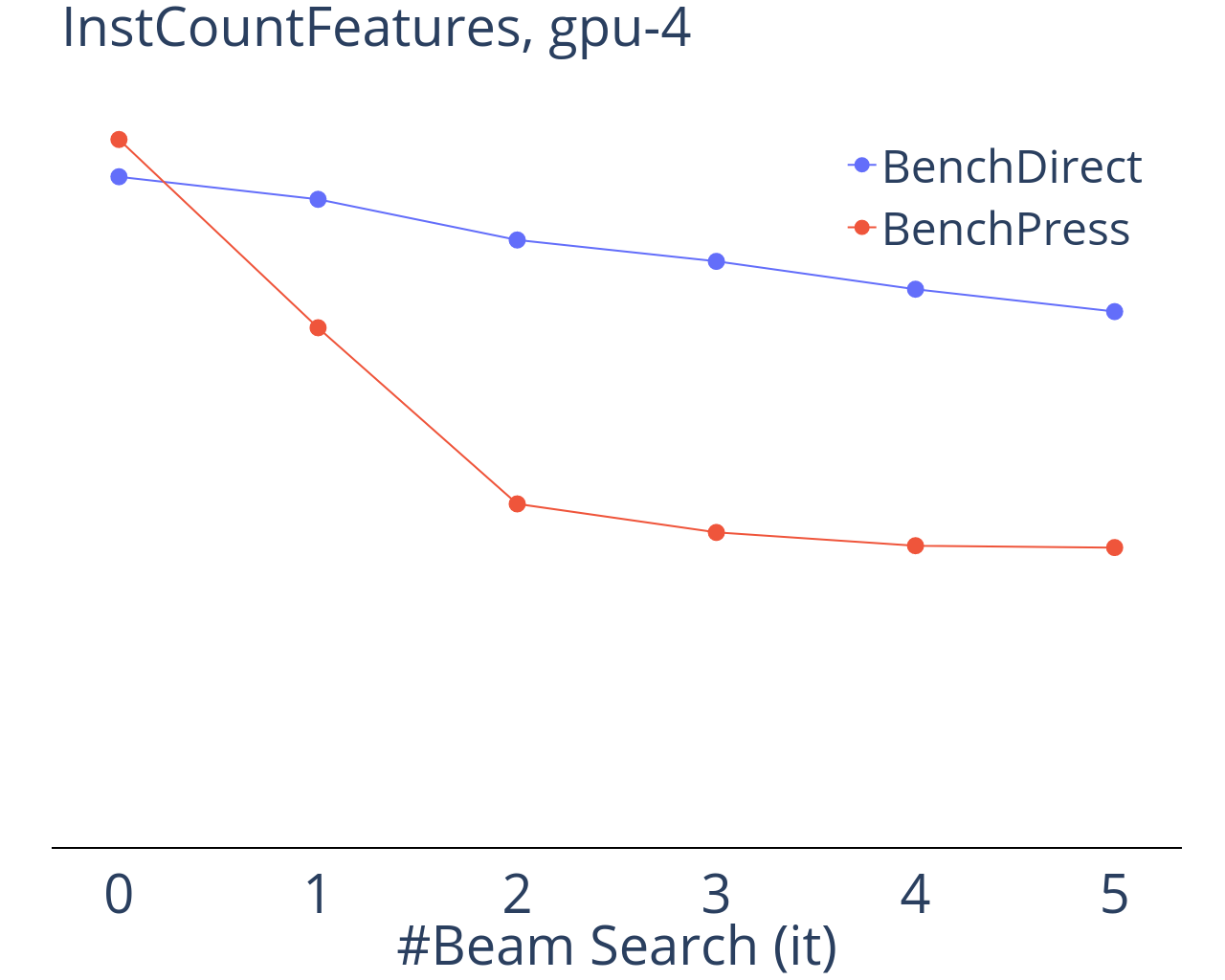}
        &
        \includegraphics[scale=0.12]{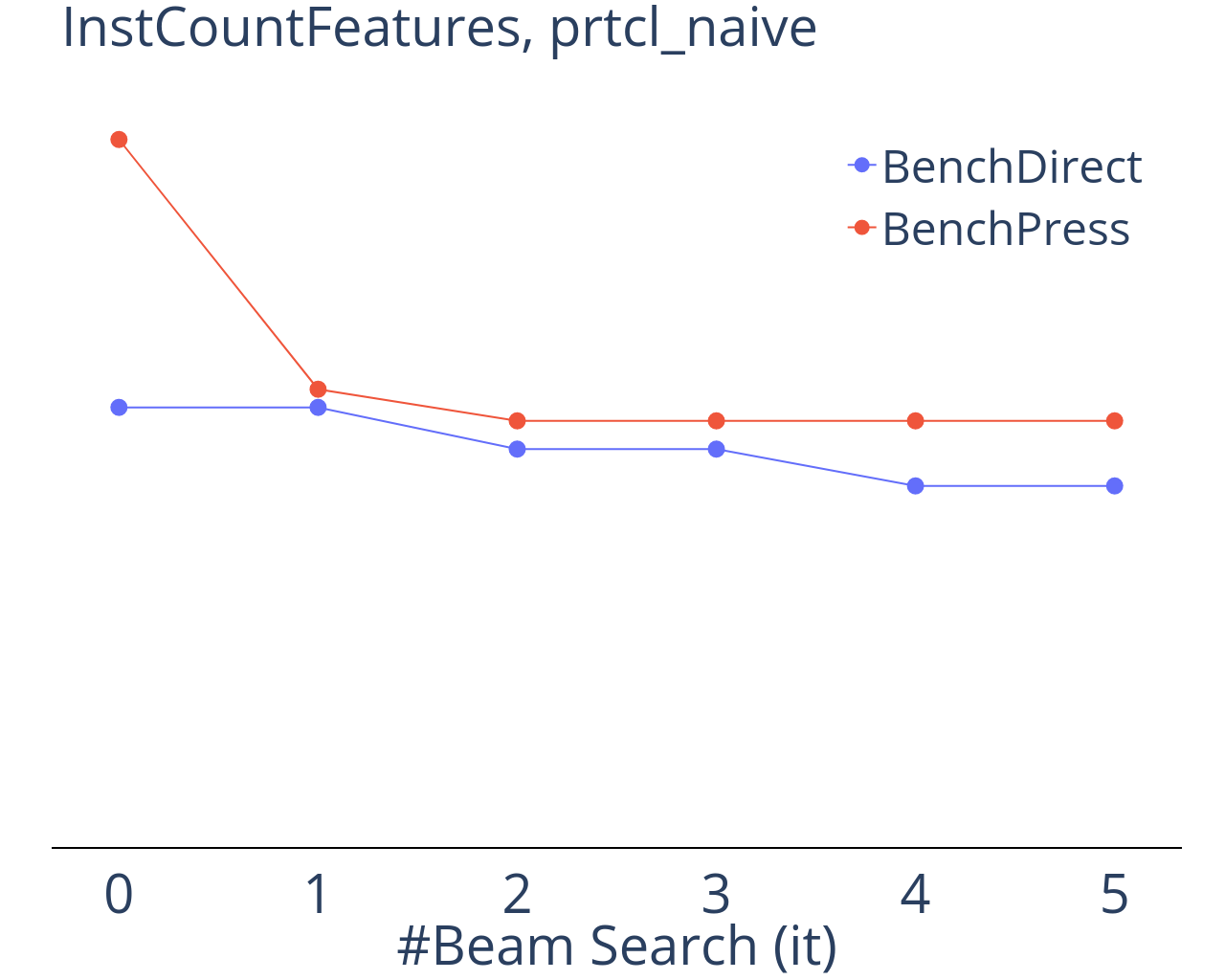}
        \\ [0.5cm]
        \includegraphics[scale=0.12]{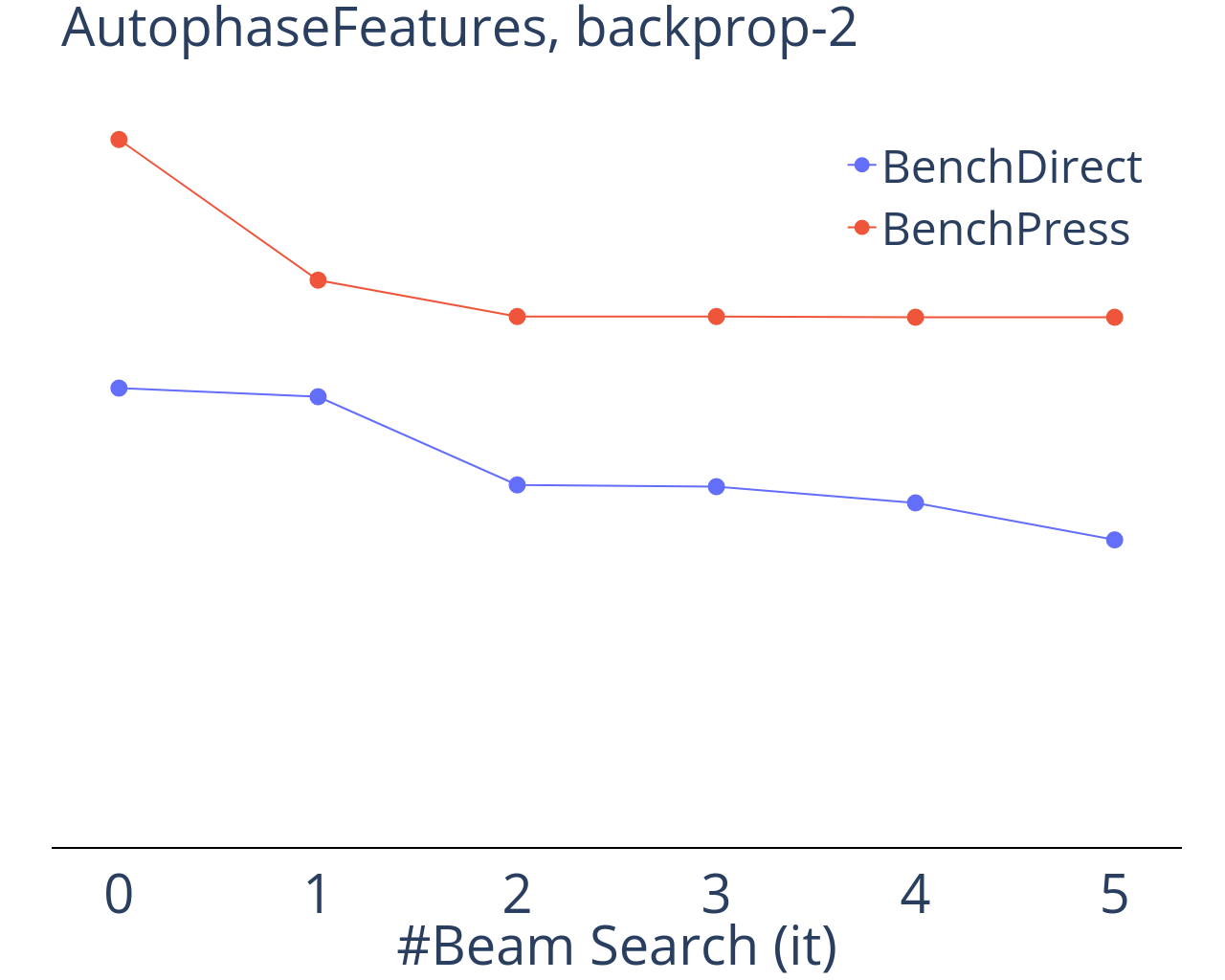}
        &
        \includegraphics[scale=0.12]{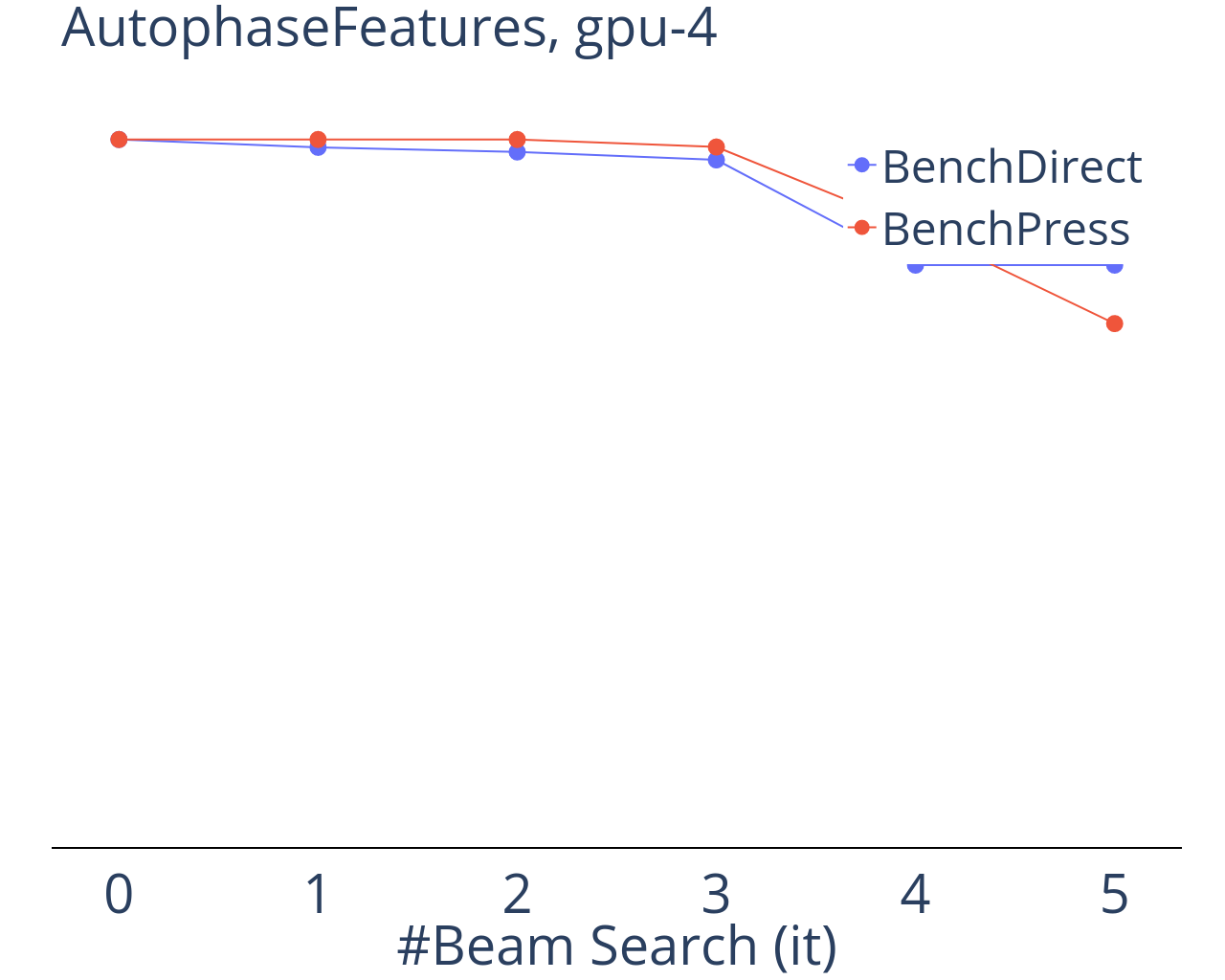}
        &
        \includegraphics[scale=0.12]{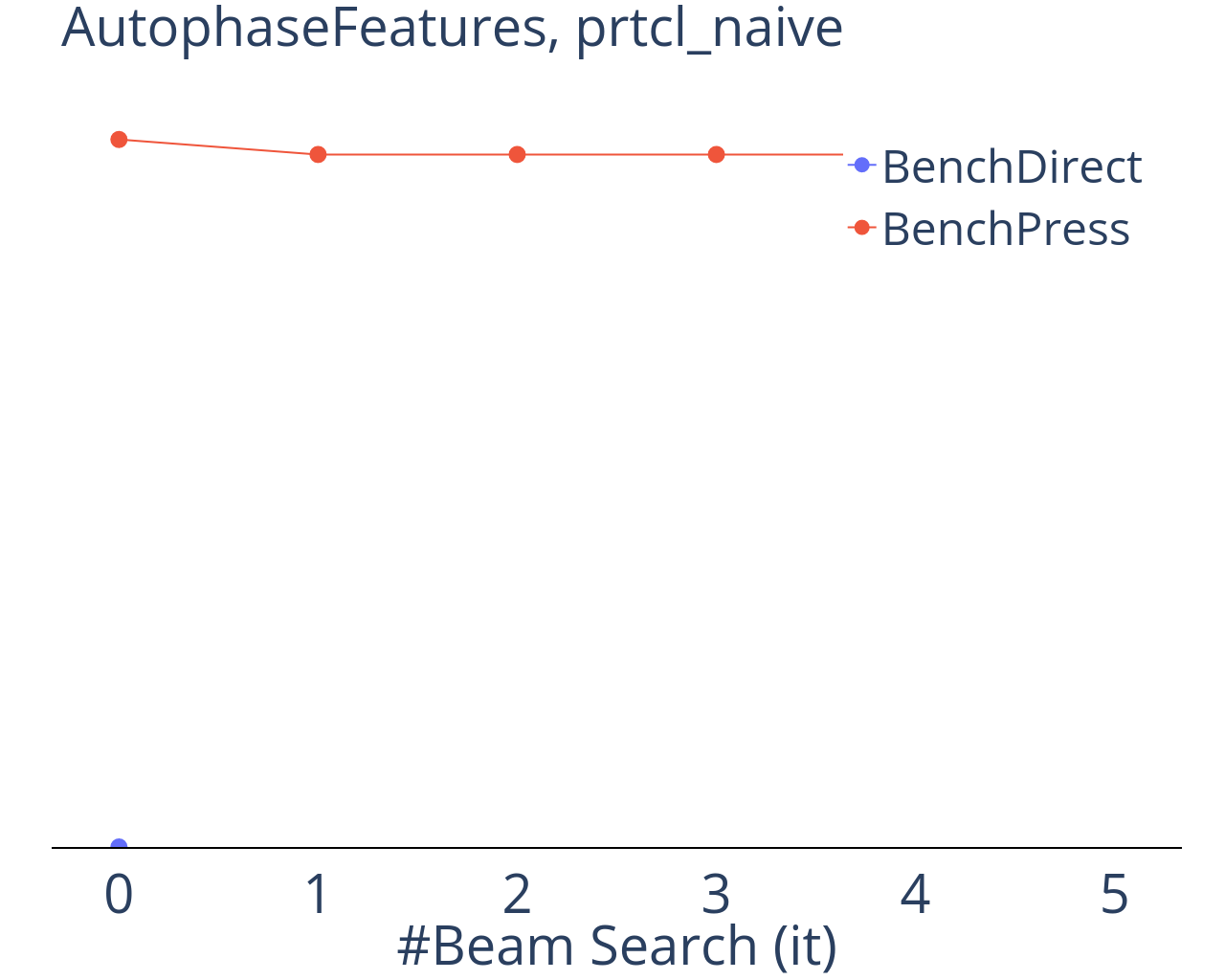}
        \\ [0.5cm]
    \end{tabular}  
    \vspace{-10pt}
    \caption{A qualitative comparison between \texttt{BenchDirect} and \name\ for \texttt{backprop-2}, \texttt{gpu-4} and \texttt{particle\_naive} Rodinia benchmarks in all three feature spaces. We show for both language models the minimum distance achieved (y-axis) from the target over the course of five beam search iterations (x-axis).}
    \vspace{-5pt}
    \label{fig:beam_search}
\end{figure*}

Both models achieve a peak accuracy when they use a workload size of 2048. This is expected as generating more candidates increases the probability of getting closer to the target features. Using this configuration on both models, we show in Figures \ref{fig:directed_search_results}a, \ref{fig:directed_search_results}b and \ref{fig:directed_search_results}c the best relative proximity achieved for each target benchmark in all three feature spaces. Similarly to Figures~\ref{fig:search_results}a, ~\ref{fig:search_results}b and ~\ref{fig:search_results}c, candidates whose euclidean distance from the target is 0 (i.e., perfect match feature-wise) are marked with a white asterisk (*). For a selection of Rodinia target benchmarks, we show how the minimum distance from the target is reduced over the course of 5 beam search iterations for both models in Figure~\ref{fig:beam_search}.

\texttt{BenchDirect} generates $1.8\times$ more candidates that match exactly the target features compared to \name. Specifically, it matches 21 targets on Grewe's et al. features, 14 on InstCount and 10 targets on Autophase, compared to \name's 17, 3 and 5 exact matches respectively. Overall, \texttt{BenchDirect} gets closer to the target compared to \name. Its samples are closer, or as close, for 45 out of 58 Rodinia targets on Grewe's et al. features, 47 out of 52 on InstCount and 49 out of 52 on Autophase. \name\ provides better candidates for 13, 5 and 3 targets on Grewe's et al., InstCount and Autophase features respectively. Even though it is expected for \texttt{BenchDirect} to miss some target features due to the experiment's randomness, we pick out a few such examples to discuss why this happens.

The largest performance gap in favour of \name\ is observed on \texttt{ellipse} and \texttt{ellipse\_opt} on InstCount features. These two benchmarks are very large, containing multiple thousands of instructions, therefore they are difficult kernels to target. We examine both models' generated samples over all 5 beam search iterations. In both cases, we find \texttt{BenchDirect}'s closest candidate on the first iteration to be 8\% closer to the target compared to \name's. After measuring the distance distribution from the target for both models' samples, we find \texttt{BenchDirect} is 93\% more likely to generate a sample whose distance is lower compared to \name\ on the first beam search iteration. \texttt{BenchDirect} seems to succeed in these two target benchmarks indeed. However, at every inference step \texttt{BenchDirect} tries to match the target features in a single \texttt{[HOLE]} infill. As these two kernels are very large, this is a challenging task leading to most of its produced candidates to have syntactic errors, leaving it with only a few benchmarks that compile. Even though its first iteration's samples are closer compared to \name, all successive iterations are becoming increasingly difficult for \texttt{BenchDirect} to produce a compiling kernel which also reduces the minimum distance. For that reason, \name's random and cautious steps lead to benchmarks that are eventually closer. We notice this pattern to happen in all targets where \name\ produced a better candidate. For these targets, it is likely that if we broke down the difficulty into smaller steps by using intermediate feature vectors, this would have helped \texttt{BenchDirect} to get to the target features gradually but more accurately.

\subsection{Human Likeness of Code}

We conduct an empirical evaluation on \name, \texttt{BenchDirect}, \texttt{CLgen} and \texttt{CLSmith} to measure the human-likeness of their samples by devising a Turing test in the form of a web application. Human-likeness is a desirable property for programs synthesized by generative models, as it indicates samples are likely to assimilate the functionality of human-written benchmarks. Each participant is shown a benchmark picked randomly from one of the 5 following datasets, (a) \name, (b) \texttt{BenchDirect}, (c) \texttt{CLgen}, (d) \texttt{CLSmith}, and (e) \texttt{GitHub}. They are then asked to label the benchmark as written by a human or an AI. During this test, we show only the benchmarks that were selected in experiments~\ref{subsec:res_target} and~\ref{subsec:res_target_direct}, i.e., the closest samples per dataset to Rodinia for all 3 feature spaces. This results in 168 samples per presented dataset.

In total, we collect data from 77 participants that declare familiarity with programming. Table~\ref{tab:human_like} shows how often users tag a test from each dataset as `human-written'. We notice that human-written code from \texttt{GitHub} is classified as `AI-written' by users in 49\% of the tests. We believe this to be due to two reasons. First, the dataset from \texttt{GitHub} contains large OpenCL kernels that contain long and unnatural expressions or have had their loops manually unrolled for optimisation reasons, making them hundreds of lines long. Such kernels are most of the times labelled as `AI-written'. Second, a participant may be suspicious of statements that do not look simple enough to be written by a human, therefore tending to select the `AI-written' label more often.


\begin{table}[]
\begin{tabular}{lccc}
           & \begin{tabular}[c]{@{}l@{}}Score \%\end{tabular} & \begin{tabular}[c]{@{}l@{}}\#Human\end{tabular} & \begin{tabular}[c]{@{}l@{}}\#Total\end{tabular}\\\hline
\texttt{GitHub} & 51\% & 139 & 270\\
\texttt{BenchPress} & 53\% & 55 & 103\\
\texttt{BenchDirect} &  49\% & 60 & 122\\
\texttt{CLgen}      & 38\% & 36 & 95\\
\texttt{CLSmith}     & 29\% & 26 & 89\\ 
\end{tabular}
\caption{Score of `human-likeness' expressed as the percentage of code examples from each dataset that were tagged as `human-written' by users}
\label{tab:human_like}
\end{table}

Participants label samples from \name\ as `human-written' in 53\% of its total tests and 49\% of \texttt{BenchDirect}'s total tests. While both scores are similar, it is likely that \texttt{BenchDirect} produces statements that are not likely written by a human slightly more often than \name. This is because it tends to generate longer sequences than \name\ when trying to reach to outliers of the feature space in a single inference step. \texttt{CLgen}'s samples may look human likely but most of them are short, no longer than 3-4 lines. Often they contain no workloads or loops and are accompanied by unused arguments. This is the reason it scores lower at 38\%. Finally, \texttt{CLSmith} is the most obvious case of unstructured and complicated code, being classified as `human-written' only 29\%. This fuzzer generates kernels by producing random expressions that conform to OpenCL's grammar, leading to random code whose functionality is not clear.

%% file: 08-Conclusion.tex
Predictive models for compilers have been shown to outperform compiler experts but they are restricted by the amount and quality of training data they are exposed to. What is needed is an approach that can synthesize benchmarks and enhance datasets with missing features. In this paper we propose \name, a powerful code generator that uses active learning to search the feature space and steers generation towards desired features. \name\ generates $10\times$ more and $7.5\times$ larger undirected benchmarks with $37\times$ greater compilation rate than \texttt{CLgen} - a state of the art compiler benchmark generator - from a fixed input feed. \name\ outperforms \texttt{CLgen}, \texttt{CLSmith}, code from \texttt{GitHub} and applied mutations with \texttt{SRCIROR} in generating OpenCL kernels that target the features of Rodinia benchmarks developed by human experts. \name\ applies active learning to enhance Grewe's et al. dataset with benchmarks with missing features and leads to improving the heuristic's speedup by 50\%. We further extend \name's language model into a directed synthesizer given compiler features. This directed model produces $1.8\times$ more matches to target features, it improves the generation process's accuracy by up to 36\% and reduces inference time by up to 72\%, while we show both our techniques outperform all other synthetic benchmark generation techniques in producing high-quality programs that are indistinguishable from human-written benchmarks. We hope this work to demonstrate a sustainable method to direct feature space search of program generation and that \name's release to researchers will enable research in related domains.

%% file: 06-Background.tex
\name\ is inspired by BERT, a representation model by Devlin et al.~\cite{bert}. Contrary to previous techniques~\cite{matthew, Radford2018ImprovingLU}, BERT learns on unlabeled text data by jointly conditioning on both left and right context. BERT enables multiple applications of this architecture to a wide variety of difficult machine learning tasks, including programming languages. In CuBERT~\cite{cubert}, Kanade et al. apply BERT over Python programs and evaluate it on finding typical mutation faults. In CodeBERT~\cite{codebert}, Feng et al. fine-tune BERT to perform NL-PL and PL-NL transformations. In this work, we extend BERT to a bidirectional generative model, with the help of \texttt{[HOLE]} token.

Cummins et al.~\cite{clgen} develop \texttt{CLgen}, a deep learning generator based on LSTM~\cite{lstm} for OpenCL programs. They try to tackle the compiler benchmark shortage by providing synthetic benchmarks as training data for compiler heuristics. The authors present the Grewe et al.~\cite{grewe} heuristic model improved its performance by $1.27\times$ when trained on their synthetic benchmarks. However, Goens et al.~\cite{goens} show that training with \texttt{CLgen}'s synthetic samples lead to a slowdown compared to training on human-written benchmarks only. To explain this, they measure the AST depth of \texttt{CLgen}'s samples and show it is $3\times$ smaller compared to human-written benchmarks and code from \texttt{GitHub} and poor in features, therefore unrealistic. This motivates us to develop \name, which produces $10\times$ more unique kernels that are $7.5\times$ larger on average.

In 2019, Nye et al. develop SketchAdapt~\cite{Nye2019LearningTI}, which uses a generator-synthesizer~\cite{deepcoder, 10.5555/3305381.3305484} to generate program sketches given I/O specifications. The synthesizer samples sketches and the generator fills <HOLE> tokens with statements. SketchAdapt performs better than other architectures~\cite{deepcoder, 10.5555/3305381.3305484}, however it samples only a pre-defined pool of operations, which restricts its diversity. Bruen et al.~\cite{debruin2021autoencoders}, propose a Tree2Tree approach for code generation using VAE. They encode AST nodes using Tree-LSTMs (Tai et al.~\cite{tai-etal-2015-improved}) and train their model on C++ functions. They test their approach against a VAE with an LSTM Seq2Seq model. They use their model as a synthesizer by sampling random AST representations which they extend to new programs. Their Seq2Seq model achieves a compilation rate of up to 67\% with greedy search, however this happens because the model greedily selects the most probable labels, leading to repetitive samples. When sampling with temperature, their Tree2Tree architecture is able to generate a wider variety of samples, but only achieves a compilation rate of 22\%, which translates to a few functions.

Gupta et al.~\cite{gupta} develop SED, a two-stage generator. A synthesizer receives I/O specifications and generates programs likely to satisfy them and a neural debugger applies program repair to reform them into functions that match specifications. Gupta et al. evaluate three synthesizer architectures and measure (a) the correctness of generated programs across tests and (b) the accuracy of their debugger to repair code. While SED is an innovative work, Karel is a small-scale language and SED's generative performance on a complex programming language is not evaluated. Faustino et al. develop Anghabench~\cite{Anghabench} to tackle the benchmark shortage~\cite{clgen, 8357388}. Anghabench is a collection of C programs mined from \texttt{GitHub}. In order to make it compilable, they use Psyche-C~\cite{psychec} type inference engine to apply type reconstruction and resolve dependencies. Structs, unions and other composite data types are omitted or re-declared with primitive types. Their benchmarks are compiling, but cannot be executed. Compared to AnghaBench, \name\ resolves type dependencies of composite types and user-defined functions without changing the functionality or semantics of programs.